%% file: NormReview.tex
\documentclass[10pt,journal,compsoc]{IEEEtran}

\usepackage{times}
\usepackage{graphicx}
\usepackage{amsmath,amssymb}
\usepackage{color}
\usepackage[pagebackref=false,breaklinks=true,colorlinks,bookmarks=false]{hyperref}
\usepackage[normalem]{ulem}
\usepackage{subfigure}
\usepackage{booktabs}
\usepackage{multirow}
\usepackage{footnote}
\usepackage{pifont}
\usepackage{xspace}
\usepackage{amsthm}

\usepackage{nicefrac}       
\usepackage{microtype}      
\usepackage{array}
\usepackage{enumitem}
\usepackage{algorithm}
\usepackage{algorithmic}

\newtheorem{defn}{Definition}[section]

\input{math_commands.tex}

\makeatletter
\DeclareRobustCommand\onedot{\futurelet\@let@token\@onedot}
\def\@onedot{\ifx\@let@token.\else.\null\fi\xspace}

\def\eg{\emph{e.g}\onedot}

\def\ie{\emph{i.e}\onedot}

\def\etc{\emph{etc}\onedot}

\def\wrt{w.r.t\onedot}

\def\etal{\emph{et al}\onedot}

\newcommand{\X}[1]{\mX_{#1}}

\newcommand{\hx}[1]{\hat{\rvx}_{#1}} 
\newcommand{\HX}[1]{\widehat{\mX}_{#1}} 
\newcommand{\HZ}[1]{\widehat{\mX}_{#1}} 
\newcommand{\TX}[1]{\widetilde{\mX}_{#1}} 

\newcommand{\TtX}[1]{\widetilde{\tX}_{#1}} 

\newcommand{\WM}[1]{\Sigma^{-\frac{1}{2}}_{#1}} 
\newcommand{\CM}[1]{\Sigma_{#1}} 

\newcommand{\Pro}{\Pi}  
\newcommand{\OP}{\Phi} 
\newcommand{\RC}{\Psi} 

\newcommand{\W}[1]{\mathbf{W}_{#1}}

\newcommand{\D}[1]{\frac{\partial \mathcal{L}}{\partial #1 }}

\newcommand{\Dl}[1]{\frac{\partial \ell}{\partial #1}}

\newcommand{\x}[1]{\rvx_{#1}}

\newcommand{\Eqn}{Eqn.}

\newcommand{\Para}[1]{\noindent\textbf{#1}}

\newcommand{\Term}[1]{\textit{#1}}

\ifCLASSOPTIONcompsoc
\usepackage[nocompress]{cite}
\else
\usepackage{cite}
\fi

\begin{document}

\title{Normalization Techniques in Training DNNs: Methodology, Analysis and Application}

\author{Lei~Huang,
	    Jie~Qin,
    Yi~Zhou,
    Fan~Zhu,
    Li~Liu,
    Ling~Shao
    \IEEEcompsocitemizethanks{
        \IEEEcompsocthanksitem  Lei~Huang, Jie~Qin, Yi~Zhou,  Fan~Zhu, Li~Liu  and Ling~Shao are with the Inception Institute of Artificial Intelligence, Abu Dhabi, UAE.}
    \thanks{Corresponding author: Lei~Huang (huanglei36060520@gmail.com).}
}

\markboth{Normalization Techniques in Training DNNs: Methodology, Analysis and Application}%
{L.~Huang, L.~Liu, F.~Zhu,  L.~Shao}
%

\IEEEtitleabstractindextext{%
\begin{abstract}
Normalization techniques are essential for accelerating the  training and improving the generalization of deep neural networks (DNNs), and have successfully been used in various applications. This paper reviews and comments on the past, present and future of normalization methods in the context of DNN training. We provide a unified picture of the main motivation behind different approaches from the perspective of optimization, and present a taxonomy for understanding the similarities and differences between them. Specifically, we decompose the pipeline of the most representative normalizing activation methods into three components: the normalization area partitioning,  normalization operation and normalization representation recovery. In doing so,  we provide insight for designing new normalization technique. Finally, we discuss the current progress in understanding normalization methods, and provide a comprehensive review of the applications of normalization for particular tasks, in which it can effectively solve the key issues.

\end{abstract}

}

\maketitle

\IEEEdisplaynontitleabstractindextext

\IEEEpeerreviewmaketitle

\IEEEraisesectionheading{\section{Introduction}
	\label{sec:introduction}}

\IEEEPARstart{D}{eep} neural networks (DNNs) have been extensively used across a broad range of applications, including computer vision (CV), natural language processing (NLP), speech and audio processing, robotics, bioinformatics, \etc~\cite{2016_book_Ian}.
They are typically composed of stacked layers/modules, the transformation between which consists of a linear mapping with learnable parameters and a nonlinear activation function~\cite{2015_Nature_LeCun}. 
While their deep and complex structure provides them powerful representation capacity and appealing advantages in learning feature hierarchies, it also makes their training difficult~\cite{2010_AISTATS_Glorot,2013_ICML_Pascanu}.
In fact, the success of DNNs heavily depends on breakthroughs in training techniques~\cite{2006_Science_Hinton,2010_ICML_Nair,2014_arxiv_Kingma,2015_ICML_Ioffe}, which has been witnessed by the history of deep learning~\cite{2016_book_Ian}.

One milestone technique in addressing the training issues of DNNs was batch normalization (BN)~\cite{2015_ICML_Ioffe}, which standardizes the activations of intermediate DNN layers  within a mini-batch of data. BN improves DNNs' training stability, optimization efficiency and generalization ability. It is a basic component in most state-of-the-art architectures~\cite{2016_CVPR_He,2015_CVPR_Szegedy,2016_BMVC_Zagoruyko,2016_CVPR_Szegedy,2016_ECCV_He,2017_CVPR_Qi,2017_CVPR_HuangGao,2017_CVPR_Xie}, and has successfully proliferated throughout various areas of deep learning~\cite{2015_ImageNet,2014_ECCV_Lin,2015_arxiv_chang}. Further, a significant number of other normalization techniques have been proposed to address the training issues in particular contexts,  further evolving the DNN architectures and their applications~\cite{2016_NeurIPSW_Ba,2016_NeurIPS_Salimans,2018_ECCV_Wu,2018_ICLR_Miyato,2020_CVPR_Huang2}. For example, layer normalization (LN)~\cite{2016_NeurIPSW_Ba} is an essential module in Transformer~\cite{2017_NeurIPS_Vaswani}, which has advanced the state-of-the-art architectures for NLP~\cite{2017_NeurIPS_Vaswani,2018_ICLR_Yu,2019_NeurIPS_Xu,2020_ICML_Xiong}, while spectral normalization~\cite{2018_ICLR_Miyato} is a basic component in the discriminator of generative adversarial networks (GANs)~\cite{2018_ICLR_Miyato,2019_ICML_Kurach,2019_ICLR_Brock}.
Importantly, the ability of most normalization techniques to stabilize and accelerate training has helped to simplify the process of designing network architectures---training is no longer the main concern, enabling more focus to be given to developing components that can effectively  encode prior/domain knowledge into the architectures.

However, despite the abundance and ever more important roles of normalization techniques, we note that there is an absence of a unifying lens with which to  describe, compare and analyze them. 
This paper provides a review and commentary on  normalization techniques in the context of training DNNs.  
To the best of our knowledge, our work is the first survey paper to cover normalization methods, analyses and applications. 
We attempt to provide answers for the following questions:

(1) What are the main motivations behind different normalization methods in DNNs, and how can we present a taxonomy for understanding the similarities and differences between a wide variety of approaches?

(2) How can we reduce the gap between the empirical success of normalization techniques  and our theoretical understanding of them?

(3) What recent advances have been made in designing/tailoring normalization techniques for different tasks, and what are the main insights behind them?

We answer the first question by providing a unified picture of the main motivations behind different normalization methods, from the perspective of optimization (Section~\ref{sec:motivaiton-overview}). We show that most normalization methods are essentially designed to satisfy  nearly equal statistical distributions of layer input/output-gradients across different layers during training, in order to avoid the ill-conditioned landscape of optimization.
  Based on this, we provide a comprehensive review of the normalization methods, including  normalizing activations by population statistics (Section~\ref{sec:normalizing-activation-population}), normalizing activations as functions (Section~\ref{sec:normalizing-function}), normalizing weights (Section~\ref{sec:normalizing-weight}) and normalizing gradients (Section~\ref{sec:normalizing-gradient}). Specifically, we decompose the most representative  normalizing-activations-as-functions framework into three components: the normalization area partitioning (NAP), normalization operation (NOP) and normalization representation recovery (NRR). We unify most normalizing-activations-as-function methods into this framework, and provide insights for designing new normalization methods.

 To answer the second question,  we discuss the recent progress in our theoretical understanding of BN in Section~\ref{sec:theory}. It is difficult to fully analyze the inner workings of BN in a unified framework, but our review ultimately provides clear guidelines for understanding why BN stabilizes and accelerates training, and further improves generalization, through a scale-invariant analysis, condition analysis and stochasticity analysis, respectively.

We answer the third question in Section~\ref{sec:application}  by providing a review of the applications of normalization for particular tasks, and illustrating how normalization methods can be used to solve key issues.
To be specific, we mainly review the applications of normalization in domain adaptation, style transfer, training GANs and efficient deep models. We show that the normalization methods can be used to `edit' the statistical properties of layer activations. These statistical properties, when designed well,  can represent the style information for a particular image or the domain-specific information for a distribution of a set of images. This characteristic of normalization methods has been thoroughly exploited in CV tasks and potentially beyond them.

We conclude the paper with additional thoughts about certain open questions in the research of normalization techniques.

\section{Denotations and Definitions}
\label{sec:denotation}
In this paper, we use a lowercase letter $x \in \R$ to denote a scalar,  boldface lowercase letter $\rvx \in \R^{d}$  for a vector,  boldface uppercase letter for a matrix $\mX \in \R^{d \times m}$, and boldface sans-serif notation for a tensor $\tX$, where $\R$ is the set of real-valued numbers, and $d, m$ are positive integers. Note that a tensor is a more general entity. Scalars, vectors and matrices can be viewed as 0th-order, 1st-order and 2nd-order tensors. Here,  $\tX$ denotes a tensor with an order larger than 2. We will provide a more precise definition in the later sections. We follow  matrix notation where the vector is in column form, except that the derivative is a row vector.

\subsection{Optimization Objective}
Consider a true data distribution $p_{*}(\rvx, \rvy) =p(\rvx) p(\rvy|\rvx)$ and the sampled training sets $\sD \sim p_{*}(\rvx, \rvy)$ of size $N$.
We focus on a supervised learning task aiming to learn the conditional distribution $p(\rvy|\rvx)$ using the model $q(\rvy|\rvx)$, where $q(\rvy|\rvx)$ is represented as a function $f_{\theta}(\rvx)$ parameterized by $\theta$.
Training the model can be viewed as tuning the parameters to minimize the discrepancy between the desired output $\rvy$ and the predicted output $f(\rvx; \theta)$. This discrepancy is usually described by a loss function $\ell(\rvy, f(\rvx; \theta))$ for each sample pair $(\rvx, \rvy)$. The empirical risk, averaged over the sample loss  in training sets $\sD$, is defined as:

\begin{equation}
\label{eqn:emprical_risk}
\mathcal{L}(\theta)=\frac{1}{N} \sum_{i=1}^{N} (\ell(\rvy^{(i)}, f_{\theta}(\rvx^{(i)}))).
\end{equation}

This paper mainly focuses on discussing the empirical risk from the perspective of optimization. We do not explicitly analyze the risk under the true data distribution $\mathcal{L}^*(\theta)= \E_{(\rvx, \rvy) \sim p_{*}(\rvx, \rvy)} (\ell(\rvy^{(i)}, f_{\theta}(\rvx^{(i)})))$ from the perspective of generalization.

\subsection{Neural Networks}
The function $f(\mathbf{x}; \theta)$ adopted by neural networks usually consists of stacked layers.
For a multilayer perceptron (MLP), $f_{\theta}(\mathbf{x})$ can be represented as a layer-wise linear and nonlinear transformation, as follows:
\begin{eqnarray}
\label{eqn:MLP}
\rvh^{l}&= &\mW^{l} \rvx^{l-1},   \\
\rvx^{l}&=& \phi(\rvh^l), ~~~ l=1,..., L,
\end{eqnarray}
where $\rvx^0 = \rvx$, $\mW^l \in \R^{d_{l} \times d_{l-1}}$ and $d_{l}$ indicates the number of neurons in the $l$-th layer. The learnable parameters $\mathbf{\theta}=\{ \mW^l, l=1,...,L \}$. Typically, $\rvh^l$ and $\rvx^l$ are  referred to as the pre-activation and activation, respectively, but in this paper, we refer to both as activations for simplicity. We also set  $\rvx^L=\rvh^L $ as the output of the network $f_{\theta}(\rvx)$ to simplify denotations.

\Para{Convolutional Layer:}
The convolutional layer parameterized by weights $\tW \in \mathbb{R}^{d_l \times d_{l-1} \times F_h \times F_w}$, where  $F_h$ and  $F_w$ are the height and width of the filter,
  takes feature maps (activations) $\tX \in \mathbb{R}^{d_{l-1} \times h \times w}$ as input, where $h$ and $w$ are the height and width of the feature maps, respectively. We denote  the set of spatial locations as $\Delta$ and the set of spatial offsets as $\Omega$. For each output feature map $k$ and its spatial location $\delta \in \Delta$, the convolutional layer computes the pre-activation $\{ \etH_{k,\delta}\}$ as: $\etH_{k,\delta}=\sum_{i=1}^{d_{l-1}} \sum_{\tau \in \Omega }  \etW_{k,i,\tau} \etX_{i, \delta+\tau}  =<\rvw_k, \rvx_{\delta}>$. Therefore, the convolution operation is a linear (dot) transformation.
 Here, $\rvw_k \in \R^{d_{l-1} \cdot F_h \cdot F_w}$  can eventually be viewed as an unrolled filter produced by $\tW$.

\subsection{Training DNNs}

From an optimization perspective, we aim to minimize the empirical risk $\mathcal{L}$, as:
  \begin{eqnarray}
\label{eqn:optim}
	\theta^* &=\arg \min_{\theta} \mathcal{L}(\theta).
\end{eqnarray}

In general, the gradient descent (GD) update is used to minimize $\mathcal{L}$,  seeking to iteratively reduce the loss as:
 \begin{eqnarray}
\label{eqn:gradientDescent}
	\mathbf{\theta}_{t+1} = \mathbf{\theta}_{t} - \eta \D{\mathbf{\theta}},
\end{eqnarray}
  where $\eta$ is the learning rate. For large-scale learning, stochastic gradient descent (SGD) is extensively used to approximate the gradients $\D{\mathbf{\theta}} $ with a mini-batch gradient.
  One essential step is to calculate the gradients. This can be done by backpropagation for calculating $\D{\rvx^{l-1}}$:
\begin{eqnarray}
\label{eqn:gradients-calculation-activation}
 \Dl{\rvx^{l-1}}&=& \Dl{\rvh^{l}} \mW^{l}, \\
  ~ \Dl{\rvh^{l-1}}&=&\Dl{\rvx^{l-1}} \phi^{'}(\rvh^{l-1}), ~l=L,  ..., 2,
\end{eqnarray}
and $\Dl{\mW^{l}}$:
\begin{equation}
\label{eqn:gradients-calculation-weight}
   \D{\mW^{l}}=  \E_{\sD}[ (\rvx^{l-1}\Dl{\rvh^{l}})^T], ~~~l=L,  ..., 1.
\end{equation}

\subsection{Normalization}

Normalization is widely used in data-preprocessing~\cite{1998_NN_LeCun,2009_TR_Alex,2016_CVPR_He}, data mining and other areas. The definition of normalization may vary among different topics. In this paper, we define normalization as a general transformation, which ensures that the transformed data has certain statistical properties. To be more specific, we provide the following formal definition.
\begin{defn}
	\label{def1:Normalization}
	\textbf{Normalization:} Given a set of data  $\sD=\{\rvx^{(i)}\}_{i=1}^{N}$, the normalization operation is a function $\OP: \rvx \longmapsto \hx{}$, which ensures that the transformed data $\widehat{\sD}= \{ \hx{}^{(i)} \}_{i=1}^{N}$ has certain statistical properties.
\end{defn}

\begin{figure}[t]
	\centering
	\begin{minipage}[c]{.88\linewidth}
		\centering
		\includegraphics[width=7.0cm]{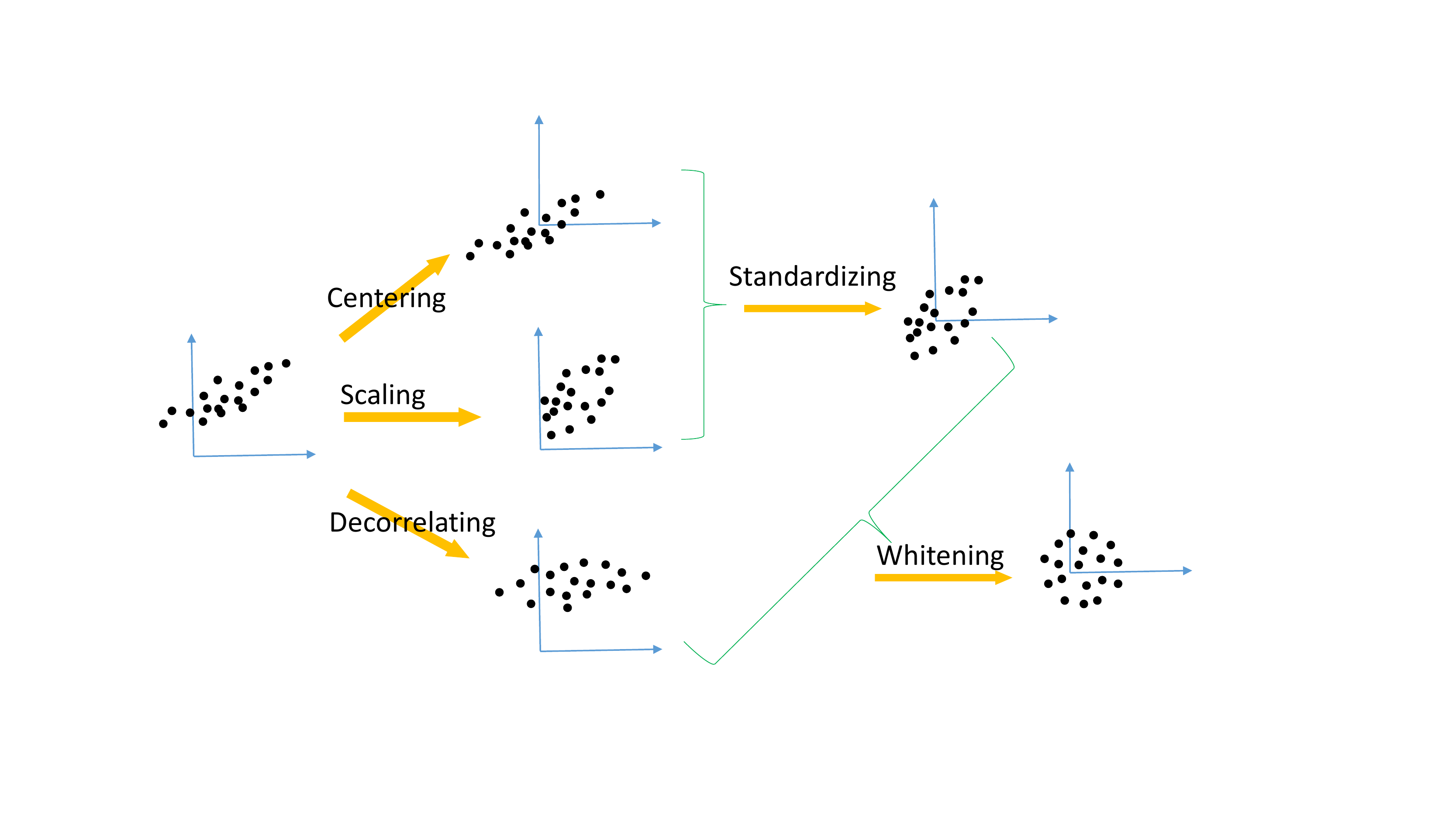}
	\end{minipage}
	\vspace{-0.06in}
	\caption{Illustration of normalization operations discussed in this paper.}
	\label{fig:NormOperation}
	\vspace{-0.12in}
\end{figure}

We consider five main normalization operations (Figure~\ref{fig:NormOperation}) in this paper: centering, scaling, decorrelating, standardizing and whitening~\cite{2018_AS_Kessy}.

\Para{Centering} formulates the transformation as:
\begin{equation}
\label{eqn-centering}
\hx{}=\OP_{C}(\rvx)
 =\rvx - \E_{\sD}(\rvx).
\end{equation}
 This ensures that the normalized output $\hx{}$ has a zero-mean property, which can be represented as: $\E_{\widehat{\sD}}(\rvx)=\mathbf{0}$.

\Para{Scaling} formulates the transformation as:
\begin{equation}
\label{eqn-scaling}
\hx{}=\OP_{SC}(\rvx)
=\Lambda^{-\frac{1}{2}} \rvx.
\end{equation}
Here,  $\Lambda=\mbox{diag}(\sigma_1^2, \ldots,\sigma_d^2)$, where $\sigma_j^2$ is the mean square over data samples for the \emph{i}-th dimension:  $\sigma_j^2 = \E_{\sD}(\rvx_j^2)$. Scaling ensures that the normalized output $\hx{}$ has a unit-variance property, which can be represented as: $\E_{\widehat{\sD}}(\hx{j}^2)=1$ for all $j=1, ..., d$.

\Para{Decorrelating} formulates the transformation as:
\begin{equation}
\label{eqn-decorrelating}
\hx{}= \OP_{D}(\rvx)
= \mD \rvx,
\end{equation}
where  $\mD=[\mathbf{d}_1, ...,
\mathbf{d}_d]$ are the eigenvectors of $\Sigma$ and $\Sigma = \E_{\sD}(\rvx \rvx^{T}) $ is the covariance matrix. Decorrelating ensures that the correlation between different dimensions of the normalized output $\hx{}$ is zero (the covariance matrix $\E_{\widehat{\sD}}(\hx{} \hx{}^{T})$ is a diagonal matrix).

\Para{Standardizing} is a composition operation that combines centering and scaling, as:
\begin{equation}
\label{eqn-standadization}
\hx{}=\OP_{ST}(\rvx)
= \Lambda^{-\frac{1}{2}} (\rvx - \E_{\sD}(\rvx)).
\end{equation}
Standardizing ensures that the normalized output $\hx{}$ has  zero-mean and unit-variance properties.

\Para{Whitening} formulates the transformation as\footnote{Whitening usually requires the input to be centered~\cite{2018_AS_Kessy,2018_CVPR_Huang}, which means it also includes the centering operation. In this paper, we unify the operation as whitening regardless of whether it includes centering or not.}:
\begin{equation}
\label{eqn-whitening}
\hx{}=\OP_{W}(\rvx)
= \tilde{\Lambda}^{-\frac{1}{2}}  \mD \rvx,
\end{equation}
where  $\tilde{\Lambda}=\mbox{diag}(\tilde{\sigma}_1, \ldots,\tilde{\sigma}_d)$ and $\mD=[\mathbf{d}_1, ...,
\mathbf{d}_d]$ are the eigenvalues and associated eigenvectors of covariance matrix $\Sigma$. Whitening ensures that the normalized output $\hx{}$ has a spherical Gaussian distribution, which can be represented as: $\E_{\sD}(\hx{} \hx{}^{T}) = \mathbf{I}$.
The whitening transformation, defined in \Eqn~\ref{eqn-whitening}, is called principal components analysis (PCA) whitening, where the whitening matrix $\WM{PCA}= \tilde{\Lambda}_{d}^{-\frac{1}{2}}  \mathbf{D}$. There are an infinite number of  whitening matrices since a whitened input stays whitened after an arbitrary rotation~\cite{2018_AS_Kessy,2020_CVPR_Huang}, which will be discussed in the subsequent sections.

\label{sec:sub-Normalization}

\section{Motivation and Overview of Normalization in DNNs}
\label{sec:motivaiton-overview}
Input normalization is extensively used in machine learning models. Intuitively, normalizing an input removes the difference in magnitude between different features, which benefits learning. There are also theoretical advantages to normalization for linear models.

 Consider a linear regression model with a scalar output $f_{\mathbf{w}}(\rvx)=\mathbf{w}^T \rvx$, and  mean square error loss $\ell=(y-f_{\theta}(\rvx))^2$.
 As shown in~\cite{1990_NeurIPS_LeCun,1998_NN_LeCun}, the learning dynamics for such a quadratic surface are fully controlled by the spectrum of the Hessian matrix $\rmH=\mathbb{E}_\sD(\rvx \rvx^T)$.
 There are two statistical momentums that are essential for evaluating the convergence behaviors of the optimization problem. One is the maximum eigenvalue of the curvature matrix  $\lambda_{max}$, and the other is the condition number of the curvature matrix, denoted by $\kappa = \frac{\lambda_{max}}{\lambda_{min}}$,
 where $\lambda_{min} $ is the  minimum nonzero eigenvalue of the curvature matrix.
 Specifically, $\lambda_{max}$ controls the upper bound and the optimal learning rate (\eg, the training will diverge if $\eta \geq  \frac{2}{\lambda_{max}(\rvh)}$). Meanwhile, $\kappa $ controls the number of iterations required for convergence (\eg, the lower bound of the iterations is $\kappa (\rvh)$ \cite{1990_NeurIPS_LeCun,1998_NN_LeCun,2018_SIAM_Bottou}). If $\rmH$ is an identity matrix that can be obtained by whitening the input through \Eqn~\ref{eqn-whitening}, the GD can converge within only one iteration. Therefore, normalizing the input can surely accelerate convergence during the optimization of linear models.

These theoretical results do not apply to neural networks directly, since the input $\rvx$ is only directly connected to the first weight matrix $\mW^{1}$, and it is not clear how input $\rvx$ affects the sub-landscapes of optimization with respect to other weight matrices $\mW^{l}, l=2, ..., L$. Luckily, the layer-wise structure (\Eqn~\ref{eqn:MLP}) of neural networks can be exploited to approximate the curvature matrix, \eg, the Fisher information matrix (FIM).
One successful example is approximating the FIM of DNNs using the Kronecker product (K-FAC) \cite{2015_ICML_Martens,2017_ICLR_Ba,2017_ICML_RelNN,2018_NeurIPS_Alberto}.
In the K-FAC approach, there are two assumptions: 1) weight-gradients in different layers are assumed to be uncorrelated; 2) the input and output-gradient in each layer are approximated as independent. Thus, the full FIM can be represented as a block diagonal matrix,  $\mathbf{F}=diag (F_1,..., F_L)$, where $F_l$ is the sub-FIM (the FIM with respect to the parameters in a  certain layer) and  computed as:
\begin{small}
	\begin{equation}
	\label{eqn:DNN_layer_FIM}
	F_{l}	\approx \mathbb{E}_{\rvx \sim p(\rvx)}[\rvx^{l-1} (\rvx^{l-1})^T] \otimes  \mathbb{E}_{(\rvx,\rvy)\sim p(\rvx)q(\mathbf{y}|\mathbf{x})}[ \Dl{\rvh^l}^T\Dl{\rvh^l}].
	\end{equation}
\end{small}
\hspace{-0.05in}Note that \cite{2014_Martens_insights,2015_ICML_Martens,2020_ECCV_Huang} have provided empirical evidence to support their effectiveness in approximating the full FIM with block diagonal sub-FIMs.
 We denote the covariance matrix of the layer input as $\Sigma_{\rvx}^l=\mathbb{E}_{p(\rvx)}[\rvx^{l-1} (\rvx^{l-1})^T]$ and the covariance matrix of the layer output-gradient as $\Sigma_{\nabla \rvh}^l =  \mathbb{E}_{q(\mathbf{y}|\mathbf{x})}[ \Dl{\rvh^l}^T\Dl{\rvh^l}]$.
Based on the K-FAC, it is clear that the conditioning of the FIM can be improved, if:
{\spaceskip=0.26em\relax
	\begin{itemize}
		\item  \emph{Criteria 1}: The statistics of the layer input (\eg, $\Sigma_{\rvx}$) and output-gradient  (\eg, $ \Sigma_{\nabla \rvh}$)  across different layers are equal.
		\item  \emph{Criteria 2}: $\Sigma_{\rvx}$ and  $ \Sigma_{\nabla \rvh}$ are well conditioned.
	\end{itemize}
}
A variety of techniques for training DNNs have  been designed to satisfy \Term{Criteria 1} and/or \Term{Criteria 2}, essentially.
For example, the weight initialization techniques aim to satisfy  \Term{Criteria 1}, obtaining nearly equal variances for layer input/output-gradients across different layers \cite{2010_AISTATS_Glorot,2015_ICCV_He,2013_CoRR_Saxe,2016_ICLR_Mishkin,2018_Arxiv_Piotr} by designing initial weight matrices. However, the equal-variance property across layers  can be broken down and is not necessarily sustained throughout training, due to the update of weight matrices.
From this perspective, it is important to normalize the activations in order to produce better-conditioned optimization landscapes, similar to the benefits of normalizing the input.

Normalizing  activations is more challenging than normalizing an input with a fixed distribution, since the distribution of layer activations $\rvx^l$ varies during training.
Besides, DNNs are usually optimized over  stochastic or mini-batch gradients, rather than the full gradient, which requires more efficient statistical estimations for activations. This paper discusses three types of  normalization methods for improving the performance of DNN training:

(1) Normalizing the activations directly to (approximately) satisfy Criteria 1 and/or Criteria 2.  Generally speaking, there are two strategies for normalizing  the activations of DNNs. One is to normalize the activations using the population statistics estimated over the distribution of activations~\cite{2012_NN_Gregoire,2014_ICASSP_Wiesler,2015_NeurIPS_Desjardins}.
The other strategy  is to normalize the activations as a function transformation, which requires backpropagation through this transformation~\cite{2015_ICML_Ioffe,2016_NeurIPSW_Ba,2018_CVPR_Huang}.

(2) Normalizing the weights with a constrained distribution, such that the activations/output-gradients (Eqns~\ref{eqn:MLP} and~\ref{eqn:gradients-calculation-activation}) can be implicitly normalized. This normalization strategy is inspired by weight initialization methods, but extends them towards satisfying the desired property during training~\cite{2016_NeurIPS_Salimans,2017_ICCV_Huang,2018_AAAI_Huang,2018_ICLR_Miyato}.

(3) Normalizing gradients to exploit the  curvature information for GD/SGD, even though the optimization landscape is ill-conditioned~\cite{2017_arxiv_Yu,2017_arxiv_You}. This involves performing normalization solely on the gradients, which may effectively remove the negative effects of the ill-conditioned landscape caused by the diversity of magnitude in gradients from different layers (\ie, Criteria 1 is not well satisfied)~\cite{2010_AISTATS_Glorot}.

\section{Normalizing Activations by Population Statistics}
\label{sec:normalizing-activation-population}
In this section, we will discuss the methods that normalize activations using the population statistics estimated over their distribution. This normalization strategy views the population statistics as constant during backpropagation.
To simplify the notation, we remove the layer index $l$ of activations $\rvx^l$ in the subsequent sections, unless otherwise stated.

Gregoire \etal~\cite{2012_NN_Gregoire} proposed to center the activations (hidden units) in a Boltzmann machine to improve the conditioning of the optimization problems, based on the insight that centered inputs improve the conditioning~\cite{1998_NN_LeCun,1998_Schraudolph,2012_AISTATS_Raiko}. Specifically, given the activation in a certain layer $\rvx$, they perform the normalization as:
\begin{equation}
\label{eqn-centering-exp}
\hx{} = \rvx - \hat{\rvu},
\end{equation}
where $\hat{\rvu}$ is the mean of activations over the training dataset.
 Note that $\hat{\rvu}$ indicates the population statistics that need to be estimated during training, and is considered as  constant during backpropagation. In ~\cite{2012_NN_Gregoire}, $\hat{\rvu}$ is estimated by running averages. Wiesler~\etal~\cite{2014_ICASSP_Wiesler} also considered centering the activations to improve the performance of DNNs, reformulating the centering normalization  by re-parameterization. This can be viewed as a pre-conditioning method.
They also used running average to estimate $\hat{\rvu}$ based on the mini-batch activations. One interesting observation in ~\cite{2014_ICASSP_Wiesler} is that the scaling operation does not yield improvements in this case. One likely reason is that the population statistics estimated by running average are not accurate, and thus cannot adequately exploit the advantages of standardization.

Desjardins \etal ~\cite{2015_NeurIPS_Desjardins} proposed to whiten the activations using the population statistics as:
\begin{equation}
\label{eqn-whitening-exp}
\hx{} =\widehat{\Sigma}^{-\frac{1}{2}} (\x{} - \hat{\rvu}),
\end{equation}
where $\widehat{\Sigma}^{-\frac{1}{2}}$ is the population statistics of the whitening matrix. One difficulty  is to accurately estimate $\widehat{\Sigma}^{-\frac{1}{2}}$. In~\cite{2015_NeurIPS_Desjardins,2017_ICML_Luo}, $\widehat{\Sigma}^{-\frac{1}{2}}$ is updated over $T$ intervals, and the whitening matrix is further pre-conditioned by one hyperparameter $\epsilon$, which balances the natural gradient (produced by the whitened activations) and the naive gradient. With these two techniques, the networks with whitened activations can be trained by finely adjusting $T$/$\epsilon$. Luo ~\cite{2017_ICML_Luo}  investigated the effectiveness of whitening the activations (pre-whitening) and pre-activations (post-whitening). They also addressed the computational issues by using online singular value decomposition (SVD) when calculating the whitening matrix.


Although several improvements in performance have been achieved, normalization by population statistics still faces some drawbacks.  The main disadvantage is the training instability, which can be caused by the inaccurate estimation of population statistics: 1) These methods usually use a limited number of data samples to estimate the population statistics, due to  computational concerns. 2) Even if  full data is available and an accurate estimation is obtained for a current iteration, the activation distribution (and thus the population statistics) will change due to the updates of the weight matrix, which is known as internal covariant shift (ICS)~\cite{2015_ICML_Ioffe}. 3) Finally, an inaccurate estimation of population statistics will be amplified as the number of  layers increases, so these methods are not suitable for large-scale networks. As pointed out in ~\cite{2015_NeurIPS_Desjardins,2017_ICML_Luo}, additional batch normalization ~\cite{2015_ICML_Ioffe} is needed to stabilize the training for large-scale networks.
%
%
%

\section{Normalizing Activations as Functions}
\label{sec:normalizing-function}

BN~\cite{2015_ICML_Ioffe} paved the way to viewing normalization statistics as functions over mini-batch inputs, and addressing backpropagation through normalization operations.
Let $x$ denote the activation for a given neuron in one layer of a DNN. BN \cite{2015_ICML_Ioffe} standardizes the neuron  within $m$ mini-batch data by:
	\begin{equation}
	\label{eqn-BN-train}
	\hat{x}^{(i)}=  \frac{x^{(i)} -u}{\sqrt{\sigma^2 +\epsilon}},	
	\end{equation}
where $\epsilon>0$ is a small number to prevent numerical instability, and $u=\frac{1}{m}  \sum_{i=1}^{m}  x^{(i)}$ and $\sigma^2 = \frac{1}{m}   \sum_{i=1}^{m} (x^{(i)}-u)^2 $ are the  mean and variance,  respectively.\footnote{Note that here $u$ and $\sigma^2$ are  functions over the mini-batch data.}
During inference, the population statistics $\{\hat{u}, \hat{\sigma}^2  \}$ are required for deterministic inference, and they are usually calculated by running average over the training iterations, as follows:
\begin{equation}
\label{eqn-BN-running_average}
\begin{cases}
\hat{u}  = (1-\lambda)  \hat{u}   + \lambda  u, \\
\hat{\sigma}^2  = (1-\lambda) \hat{\sigma}^2  + \lambda \sigma^2 .
\end{cases}
\end{equation}

Compared to the normalization methods based on population statistics, introduced in Section~\ref{sec:normalizing-activation-population}, this normalization strategy provides several advantages: 1) It avoids using the population statistics to normalize the activations, thus avoiding the instability caused by  inaccurate estimations. 2) The normalized output for each mini-batch has a zero-mean and unit-variance constraint that stabilizes the distribution of the activations, and thus benefits training. For more discussions please refer to the subsequent  Section~\ref{sec:theory}.

Due to the  constraints introduced by standardization, BN also uses an additional learnable scale parameter $\gamma \in \R$ and shift parameter $\beta \in \R$ to recover a possible reduced representation capacity~\cite{2015_ICML_Ioffe}:
\begin{equation}
\label{eqn-BN-recover}
\tilde{x} = \gamma \hat{x}  + \beta.
\end{equation}
In this paper, we also refer to the scale parameter  and shift parameter  as affine parameters.
BN has been shown to be a milestone in the deep learning community~\cite{2016_CVPR_He,2016_CVPR_Szegedy,2018_ECCV_Wu}. It is widely used in different networks~\cite{2016_CVPR_He,2015_CVPR_Szegedy,2015_ICLR_Simonyan,2016_BMVC_Zagoruyko,2016_CVPR_Szegedy,2016_ECCV_He,2017_CVPR_HuangGao,2017_CVPR_Xie} and various applications~\cite{2020_ICML_John}. However, despite its great success in deep learning, BN still faces several issues in particular contexts: 1) The inconsistent operation of BN between training and inference limits its usage in complex networks (\eg recurrent neural networks (RNNs)~\cite{2017_ICLR_Cooijmans,2016_ICASSP_Laurent,2016_NeurIPSW_Ba}) and tasks~\cite{2016_NeurIPS_Salimans2,2019_ICML_Kurach,2020_arxiv_Bhatt}; 2) BN suffers from the small-batch-size problem --- its error increases rapidly as the batch size becomes smaller~\cite{2018_ECCV_Wu}. To address BN's weaknesses and further extend its functionality, plenty of works related to feature normalization have been proposed.

In the following sections, we first propose a framework to describe normalizing-activations-as-function methods in Algorithm~\ref{alg_forward}, and review the basic single-mode normalization methods, which ensure that the normalized output has a single-mode (Gaussian) distribution.
We then introduce the approaches that extend single-mode method to multiple modes,  and that further combine different normalization methods.
 Lastly, we discuss the  more robust estimation methods that  address the small-batch-size problem of BN.

\begin{algorithm}[tb]
	\algsetup{linenosize=\footnotesize}
	\footnotesize
	\caption{Framework of algorithms normalizing activations as functions.}
	\label{alg_forward}
	\begin{algorithmic}[1]
		\begin{small}
			\STATE \textbf{Input}: mini-batch inputs $ \tX \in \mathbb{R}^{d \times m \times h \times w} $.
			\STATE \textbf{Output}: $ \TtX{} \in \mathbb{R}^{d \times m  \times h \times w}$.
			\STATE Normalization area partitioning: $\mX= \Pro (\tX)$.
			\STATE Normalization operation: $\HX{}= \Phi (\mX)$.
			\STATE Normalization representation recovery: $\TX{} = \RC(\HX{})$.
			\STATE Reshape back: $\TtX{}= \Pro^{-1} (\TX{})$.
			%
		\end{small}
	\end{algorithmic}
\end{algorithm}

\subsection{A Framework for Decomposing Normalization}
We divide the normalizing-activations-as-function framework into three abstract processes: normalization area partitioning (NAP), normalization operation (NOP), and normalization representation recovery (NRR).
We consider the more general mini-batch (of size $m$) activations in a convolutional layer $\tX \in \R^{d \times m \times h \times w}$, where $d$, $h$ and $w$ are the channel number,  height and width of the feature maps, respectively.\footnote{Note that the convolutional activation is reduced to the MLP activation, when setting $h=w=1$.}
NAP transforms the activations $\tX$ into $\mX \in \mathbb{R}^{S_1 \times S_2}$\footnote{NAP can be implemented by the reshape operation of PyTorch~\cite{2017_NeurIPS_pyTorch} or Tensorflow~\cite{2016_Tensorflow}.}, where $S_2$ indexes the set of samples used to compute the estimators.
NOP denotes the specific normalization operation (see main operations in Section~\ref{sec:denotation}) on the transformed data $\mX$. NRR is used to recover the possible reduced representation capacity.

Take BN as an example.  BN~\cite{2015_ICML_Ioffe} considers each spatial position in a feature map as a sample~\cite{2015_ICML_Ioffe,2016_JMLR_Gulcehre} and the NAP is:
\begin{equation}
\label{eqn-NAP-BN}
\X{}=\Pro_{BN}(\tX ) \in \mathbb{R}^{d \times mhw},
\end{equation}
 which means that the statistics are calculated along the batch, height, and width dimensions.
The NOP is the standardization, represented in the form of a matrix as:
\begin{equation}
\label{eqn-scaling-BN}
\HZ{}=\OP_{SD}(\X{})
= \Lambda^{-\frac{1}{2}} (\X{} - \rvu  \mathbf{1}^T).
\end{equation}
Here, $\rvu $ is the mean of data samples, $\mathbf{1}$ is a column vector of all ones, and  $\Lambda_{d}=\mbox{diag}(\sigma_1^2, \ldots,\sigma_d^2) + \epsilon \mI$, where $\sigma_j^2$ is the variance over data samples for the \emph{j}-th neuron/channel.
 The NRR is the affine transformation with channel-wise learnable affine parameters  $\gamma, \beta \in \R^{d}$, defined as:
 \begin{equation}
 \label{eqn-BN-recover-matrix}
 \TX{} = \RC_{AF}(\HX{})= \HX{} \odot (\gamma \mathbf{1}^T) + (\beta \mathbf{1}^T).
 \end{equation}

In the following sections, we will discuss the research progress along these three lines.

\subsubsection{Normalization Area Partitioning}
\label{sec:normalizing-function-area}
In this section, the default NOP is the standardization operation (\Eqn~\ref{eqn-scaling-BN}), and the NRR is the affine transform (\Eqn~\ref{eqn-BN-recover-matrix}).


LN~\cite{2016_NeurIPSW_Ba} proposes to standardize the layer input within the neurons for each training sample, to avoid the drawbacks of normalization along batch dimensions. Specifically, the NAP of LN is $\X{}=\Pro_{LN}(\tX{} ) \in \mathbb{R}^{m \times dhw}$, where the normalization statistics are calculated along the channel, height and width dimensions.  LN has the same formulation during training and inference, and is extensively used in NLP tasks~\cite{2017_NeurIPS_Vaswani,2018_ICLR_Yu,2019_NeurIPS_Xu}. 

Group normalization (GN)~\cite{2018_ECCV_Wu} generalizes LN, dividing the neurons into groups and standardizing the layer input within the neurons of each group for each sample independently. Specifically, the NAP of GN is $\X{}=\Pro_{GN}(\tX{} ) \in \mathbb{R}^{mg \times shw}$, where $g$ is the group number and $d=gs$. LN is clearly a special case of GN with $g=1$. By changing the group number $g$, GN is more flexible than LN, enabling it to achieve good performance on visual tasks limited to small-batch-size training (\eg, object detection and segmentation~\cite{2018_ECCV_Wu}).

Instance normalization (IN)~\cite{2016_arxiv_Ulyanov} proposes to normalize each single image to remove instance-specific contrast information. Specifically, the NAP of IN is $\X{}=\Pro_{IN}(\tX{} ) \in \mathbb{R}^{md \times hw}$. Due to its ability to remove style information from the inputs, IN is widely used in image style transfer tasks~\cite{2017_ICLR_Dumoulin,2018_ECCV_Huang,2017_ICCV_HuangXun}.

Position normalization (PN)~\cite{2019_NeurIPS_Li} standardizes the activations at each position independently across the channels. The NAP of PN is $\X{}=\Pro_{PN}(\tX{} ) \in \mathbb{R}^{mhw \times d}$. PN is designed to deal with spatial information, and has the potential to enhance the performance of generative models~\cite{2019_NeurIPS_Li,2018_ICLR_Karras}.

Batch group normalization (BGN)~\cite{2020_ICLR_Summers} expands the grouping mechanism of GN from being over only channels to being over both channels and batch dimensions. The NAP of BGN is  $\X{}=\Pro_{BGN}(\tX{} ) \in \mathbb{R}^{g_m g \times s_m shw}$, where  $m=g_m s_m$. BGN also normalizes over batch dimensions and needs to estimate the population statistics, similar to BN in \Eqn~\ref{eqn-BN-running_average}. However, the group mechanism  adds `examples' for normalization, thus relieving the small-batch problem of BN to some degree.

\Para{Local Normalization:} In the above normalization methods, the statistics are shared by all examples/positions in the same area that are used to calculate these statistics. There are also methods in which the  statistics for each example/position are calculated over the neighboring regions and thus vary. This kind of normalization is called
\Term{local normalization}~\cite{2019_arxiv_Ortiz}.  Jarrett \etal proposed local contrast normalization (LCN) to standardize each example's feature using the statistics calculated by its neighbors in a window of size $9 \times 9$.  Local response normalization (LRN)~\cite{2012_NeurIPS_Krizhevsky} proposes to scale the activation across $n$ `adjacent' kernel maps at the same spatial position.
Local contex normalization~\cite{2019_arxiv_Ortiz} extends the neighborhood partition, where the normalization is performed within a window of size $p \times q$, for groups of filters with a size predefined by the number of channels per group ($c\_groups$) along the channel axis. Divisive normalization (DN)~\cite{2017_ICLR_Ren}  generalizes the neighborhood partition of these local normalization methods as the choices of the summation and suppression fields~\cite{2008_CVPR_Lyu,2017_ICLR_Ren}.

\subsubsection{Normalization Operation}

As previously discussed, current normalization methods usually use a standardization operation. However, other operations can also be used to normalize the data. We divide these operations into three categories: 1) Extending standardization towards the whitening operation, which is a more general operation; 2) Variations of standardization; 3) Reduced standardizations that use only centering or scaling for some special situations.
Unless otherwise stated, the NAP is $\Pro_{BN}$, the data transferred after the NAP is denoted as $\mX \in \R^{d \times m}$, and the NRR is the affine transform as shown in \Eqn~\ref{eqn-BN-recover-matrix}.

\vspace{0.05in}
\Para{Beyond Standardization Towards Whitening:}
Huang \etal proposed decorrelated BN ~\cite{2018_CVPR_Huang}, which extends BN to batch whitening (BW). The NOP of BW is whitening, represented as:
\begin{equation}
\label{eqn-whitening-miniBatch}
\HZ{}=\OP_{W}(\X{})
= \Sigma^{-\frac{1}{2}}(\X{}-\rvu  \mathbf{1}^T).
\end{equation}
Here, $\Sigma^{-\frac{1}{2}}$ is the whitening matrix, which is calculated from  the corresponding mini-batch covariance matrix $\Sigma = \frac{1}{m} (\X{} - \rvu
\mathbf{1}^T) (\X{} - \rvu \mathbf{1}^T)^T +
\epsilon \mI$.

One main challenge for extending standardization to whitening is how to back-propagate through the inverse square root of a matrix (\ie. $\partial{\Sigma^{-\frac{1}{2}}} / \partial{\Sigma}$)~\cite{2015_ICML_Ioffe,2018_CVPR_Huang}. This can be achieved by using matrix differential calculus~\cite{2015_ICCV_Ionescu}, as proposed in~\cite{2018_CVPR_Huang}. One interesting question is the choice of how to compute the whitening matrix $\Sigma^{-\frac{1}{2}}$.
 PCA based BW with $\WM{PCA}=\tilde{\Lambda}^{-\frac{1}{2}}  \mD$ suffers significant instability in training DNNs and hardly converges, due to the so-called stochastic axis swapping (SAS), as explained in \cite{2018_CVPR_Huang}.
 Zero-phase component analysis (ZCA) whitening, using $\WM{ZCA}=\mD \tilde{\Lambda}^{-\frac{1}{2}} \mD^T$, is advocated for in~\cite{2018_CVPR_Huang}, where the  PCA-whitened input is rotated back by the corresponding rotation matrix $\mD$. ZCA whitening has been shown to avoid the SAS issue and achieve better performance over standardization (used in BN) on discriminative classification tasks  \cite{2018_CVPR_Huang}.
Siarohin \etal~\cite{2019_ICLR_Siarohin} used Cholesky decomposition (CD) based whitening  $\WM{CD}=\mL^{-1}$,  where $\mL$ is a lower triangular matrix from the CD, with $\mL \mL^T=\CM{}$. CD whitening has been shown to achieve  state-of-the-art performance in training GANs. For more details on comparing different whitening methods for training DNNs, please refer to~\cite{2020_CVPR_Huang}.


Given a particular batch size, BW may not have enough samples to obtain a suitable estimate for the full covariance matrix, which can heavily harm the performance. Group-based BW---where features are divided into groups and whitening is performed
within each one---was proposed~\cite{2018_CVPR_Huang,2020_ICLR_Ye} to control the extent of the whitening. One interesting property is that group-based BW reduces to BN if the channel number in each group is set to $1$. Besides, group-based BW also has the added benefit of reducing the computational cost of whitening. Later, Huang \etal proposed iterative normalization (IterNorm)~\cite{2019_CVPR_Huang}  to improve the computational efficiency and numerical stability of ZCA whitening, since it can avoid eigen-decomposition or SVD by employing Newton's iteration for approximating the whitening matrix $\WM{}$. An similar idea was also used in~\cite{2020_ICLR_Ye} by coupled Newton-Schulz iterations~\cite{2008_book_Higham} for whitening. One interesting property of IterNorm is that it stretches the dimensions along the eigenvectors progressively, so that the associated eigenvalues converge to 1 after normalization. Therefore, IterNorm can effectively control the extent of whitening by its iteration number.

There also exist works that impose extra penalties on the loss function to obtain approximately whitened activations~\cite{2016_ICLR_Cogswell,2016_ICDM_Xiong,2018_NeurIPS_Littwin,2020_ICLR_Joo,2020_arxiv_Zhou}, or exploit the whitening operation to improve the network's generalization~\cite{2020_arxiv_Shao,2020_arxiv_Chen}.


%
%
%
%
%
%

\vspace{0.05in}
\Para{Variations of Standardization:}
There are several variations of the standardization operation for normalizing the activations. As an alternative to the  $L^2$ normalization used to control the activation scale in a BN layer~\cite{2015_ICML_Ioffe}, the  $L^1$ normalization was proposed in~\cite{2016_arxiv_Liao,2018_arxiv_Wu,2018_NeurIPS_Hoffer} for standardization. Specifically, the dimension-wise standardization deviation of  $L^1$ normalization is: $\sigma=\frac{1}{m}\sum_{i=1}^{m} |x^{(i)} - u|$. Note that $L^2$ normalization is made equivalent to  $L^1$ normalization (under mild assumptions) by multiplying it by a scaling factor $\sqrt{\frac{\pi}{2}}$~\cite{2018_arxiv_Wu,2018_NeurIPS_Hoffer}.

 $L^1$ normalization can improve numerical stability  in a low-precision implementation, as well as provide computational and memory benefits, over  $L^2$ normalization. The merits of $L^1$ normalization originate from the fact that it avoids the costly square and root operations of $L^2$ normalization. Specifically, Wu \etal~\cite{2018_arxiv_Wu} showed that the proposed sign and absolute operations in $L^1$ normalization can achieve a $1.5\times$ speedup and reduce the power consumption by  $50\%$ on an FPGA platform.  Similar merits also exist when using $L^{\infty}$, as discussed in~\cite{2018_NeurIPS_Hoffer}, where the standardization deviation is: $\sigma=\max_i |x^{(i)}|$. The more generalized $L^p$ was investigated in~\cite{2016_arxiv_Liao} and ~\cite{2018_NeurIPS_Hoffer}, where the standardization deviation is: $\sigma=\frac{1}{m} \sqrt[p]{\sum_{i=1}^{m} (x^{(i)})^p}$.

 Yuan \etal~\cite{2019_AAAI_Yuan} proposed  generalized batch normalization (GBN), in which the mean statistics for centering and the deviation measures for scaling are more general operations, the choice of which can be guided by the risk theory. They provided some optional asymmetric deviation measures for networks with ReLU, such as,  the Right Semi-Deviation (RSD)~\cite{2019_AAAI_Yuan}. 

%
%
%
%
%
%
%

\vspace{0.05in}
\Para{Reduced Standardization:}
As stated in Section~\ref{sec:denotation}, the standardizing operation usually includes centering and scaling.
However, some works only consider one or the other, for specific situations. Note that either centering or scaling along the batch dimension can benefit optimization, as shown in~\cite{2015_ICML_Ioffe,1998_NN_LeCun}. Besides, the scaling operation is important for scale-invariant learning, and has been shown useful for adaptively adjusting the learning rate to stabilize training~\cite{2019_ICLR_Arora}.

Salimans \etal proposed mean-only batch normalization (MoBN)~\cite{2016_NeurIPS_Salimans}, which only performs centering along the batch dimension, and works well when combined with weight normalization~\cite{2016_NeurIPS_Salimans}. Yan \etal~\cite{2020_ICLR_Yan} proposed to perform scaling only in BN for small-batch-size training, which also works well when combined with weight centralization.  Shen \etal~\cite{2020_ICML_Shen} also proposed to perform scaling only in BN to improve the performance  for NLP tasks.

Karras \etal~\cite{2018_ICLR_Karras} proposed pixel normalization, where the scaling only operation is performed along the channel dimension for each position of each image. This works like PN~\cite{2019_NeurIPS_Li} but only uses the scaling operation. Pixel normalization works well for GANs~\cite{2019_NeurIPS_Li} when used in the generator.

Zhang \etal~\cite{2019_NeurIPS_Zhang } hypothesized that the re-centering invariance produced by centering in LN~\cite{2016_NeurIPSW_Ba} is dispensable and proposed to perform scaling only for LN, which is referred to as root mean square layer normalization (RMSLN). RMSLN takes into account the importance of the scale-invariant property for LN. RMSLN works as well as LN on NLP tasks but reduces the running time~\cite{2019_NeurIPS_Zhang }. This idea was also used in the variance-only layer normalization for the click-through rate (CTR)  prediction task~\cite{2020_arxiv_Wang}.
Chiley \etal~\cite{2019_NeurIPS_Chiley } also proposed to perform scaling along the channel dimension (like RMSLN) in the proposed online normalization to stabilize the training.





Singh and Krishnan proposed filter response normalization (FRN)~\cite{2020_CVPR_Singh }, which performs the scaling-only operation along each channel (filter response) for each sample independently. This is similar to IN~\cite{2016_arxiv_Ulyanov} but without performing the centering operation. The motivation is that the benefits of centering for normalization schemes that are batch independent are not really justified.

\subsubsection{Normalization Representation Recovery}
Normalization constrains the distribution of the activations, which can  benefit optimization. However, these constraints may hamper the representation ability, so  an additional affine transformation is usually used to recover the possible representation, as shown in \Eqn~\ref{eqn-BN-recover-matrix}. There are also other options for constructing the NRR.

Siarohin \etal~\cite{2019_ICLR_Siarohin} proposed a coloring transformation to recover the possible loss in representation ability caused by the whitening operation, which is formulated as:
\begin{equation}
\label{eqn-BN-coloring}
\TX{} = \RC_{LR}(\HX{}) = \HX{} \W{} + (\beta \mathbf{1}^T),
\end{equation}
where $\W{}$ is a $d \times d$ learnable matrix. The coloring transformation can be viewed as a linear layer in neural networks.

In \Eqn~\ref{eqn-BN-recover-matrix}, the NRR parameters are both learnable  through backpropagation. Several works have attempted to generalize these parameters by using a hypernetwork to dynamically generate them, which is formulated as:

\begin{equation}
\label{eqn-BN-recover-generate}
\TX{} = \RC_{DC}(\HX{})= \HX{} \odot  \Gamma_{\phi^{\gamma}} + B_{\phi^{\beta}},
\end{equation}
where $\Gamma_{\phi^{\gamma}} \in \mathbb{R}^{d \times m}$ and $ B_{\phi^{\beta}} \in \mathbb{R}^{d \times m}$ are generated by the subnetworks $\phi^{\gamma}_{\theta_{\gamma}}(\cdot )$ and $\phi^{\beta}_{\theta_{\beta}}(\cdot )$, respectively. The  affine parameters generated depend on the original inputs themselves, making them different from the affine parameters shown in \Eqn~\ref{eqn-BN-recover-matrix}, which are learnable by backpropagation.

Kim \etal proposed dynamic layer normalization (DLN)~\cite{2017_InterSpeech_Kim} in a  long short-term memory (LSTM) architecture for speech recognition, where  $\phi^{\gamma}_{\theta_{\gamma}}(\cdot )$ and  $\phi^{\beta}_{\theta_{\beta}}(\cdot )$ are separate utterance-level feature extractor subnetworks, which are
jointly trained with the main acoustic model. The input of the subnetworks is the output of the corresponding hidden layer of the LSTM. Similar ideas are also used in adaptive instance normalization (AdaIN)~\cite{2018_ECCV_Huang} and adaptive layer-instance normalization (AdaLIN)~\cite{2020_ICLR_Kim} for unsupervised image-to-image translation, where the subnetworks are MLPs and the inputs of the subnetworks are the embedding features produced by one encoder. Jia \etal~\cite{2019_CVPR_Jia} proposed instance-level meta normalization (ILMN), which utilizes an encoder-decoder
subnetwork to generate affine parameters, given the instance-level mean and variance as input. Besides, ILMN also combines the learnable affine  parameters shown in  \Eqn~\ref{eqn-BN-recover-matrix}.
Rather than using the channel-wise
affine parameters shared across spatial positions, spatially adaptive denormalization (SPADE) uses the spatially dependent $\beta,\gamma \in \mathbb{R}^{d \times h \times w}$, which are dynamically generated by a two-layer CNN with raw images as inputs.

The mechanism for generating affine parameter $\Gamma_{\phi^{\gamma}}$  shown in \Eqn~\ref{eqn-BN-recover-generate} resembles the squeeze-excitation (SE) block~\cite{2018_CVPR_Hu}, when the input of the subnetwork $\phi^{\gamma}_{\theta_{\gamma}}(\cdot )$ is $\X{}$ itself, \ie, $\Gamma_{\phi^{\gamma}}=\phi^{\gamma}_{\theta_{\gamma}}(\X{} )$. Inspired by this, Liang \etal proposed instance enhancement batch normalization (IEBN))~\cite{2020_AAAI_Liang}, which combines the channel-wise affine parameters in \Eqn~\ref{eqn-BN-recover-matrix} and the instance-specific channel-wise affine parameters in \Eqn~\ref{eqn-BN-recover-generate} using  SE-like subnetworks, with fewer parameters. IEBN can effectively regulate noise by introducing instance-specific information for BN.  This idea is further generalized by attentive normalization~\cite{2019_arxiv_Li},  proposed by Li \etal, where the affine parameters are modeled by $K$ mixture components.

Rather than using a subnetwork to generate the affine parameters, Xu \etal proposed adaptive normalization (AdaNorm)~\cite{2019_NeurIPS_Xu}, where the affine parameters  depend on the standardized output $\HX{}$ of layer normalization:
\begin{equation}
\label{eqn-BN-recover-condition}
\TX{} =  \HX{} \odot  \phi(\HX{}).
\end{equation}
Here, $ \phi(\HX{})$ used in~\cite{2019_NeurIPS_Xu} is:  $\phi(\hx{})=C(1-k\hx{})$, $C=\frac{1}{d}\sum_{i=1}^{d} \hx{i}$ and $k$ is a constant that satisfies certain constraints. Note that $\phi(\HX{})$ is treated as a changing constant (not a function) and the gradient of $\phi(\HX{})$ is detached in the implementation ~\cite{2019_NeurIPS_Xu}.
We also note that the side information can be injected into the NRR operation for conditional generative models. The typical works are conditional BN (CBN)~\cite{2017_NeurIPS_Vries} and conditional IN (CIN)~\cite{2017_ICLR_Dumoulin}. We will elaborate on these in Section~\ref{sec:application} when we discuss applications of normalization.

In summary, we list the main single-mode normalization methods under our proposed framework in Table~\ref{table-methods}.
%

\begin{table*}[th]
	\centering
			\caption{Summary of the main single-mode normalization methods, based on our proposed framework for describing normalizing-activations-as-functions methods. The order is based on the time of publication. }
			\label{table-methods}
	\begin{small}
       \begin{tabular}{p{1.66in}<{\centering}| p{1.5in}<{\centering} | p{1.1in}<{\centering} | p{1.2in}<{\centering} | p{0.8in}<{\centering}}
			\bottomrule[1pt]
			Method   & NAP  & NOP  & NRR  & Published In \\
			\hline
Local contrast normalization~\cite{2009_ICCV_Jarrett}  & $\Pro_{IN}(\tX{} ) \in \mathbb{R}^{md \times hw} $: local spatial positions& Standardizing   &  No & ICCV, 2009   \\
			\hline
Local response normalization~\cite{2012_NeurIPS_Krizhevsky}   & $\Pro_{PN}(\tX{} ) \in \mathbb{R}^{mhw \times d}$: local channels& Scaling   &  No &  NeurIPS, 2012   \\
			\hline
Batch normalization (BN)~\cite{2015_ICML_Ioffe}   & $\Pro_{BN}(\tX{} ) \in \mathbb{R}^{d \times mhw}$& Standardizing   &   Learnable $\gamma,\beta \in \R^{d}$  & ICML, 2015   \\
	\hline
Mean-only BN~\cite{2016_NeurIPS_Salimans}  & $\Pro_{BN}(\tX{})\in \mathbb{R}^{d \times mhw} $& Centering   &  No & NeurIPS, 2016   \\
	\hline
Layer normalization (LN)~\cite{2016_NeurIPSW_Ba}   & $\Pro_{LN}(\tX{} ) \in \mathbb{R}^{m \times dhw}$& Standardizing   &   Learnable $\gamma,\beta \in \R^{d}$  & Arxiv, 2016   \\
	\hline
Instance normalization (IN)~\cite{2016_arxiv_Ulyanov}   & $\Pro_{IN}(\tX{} ) \in \mathbb{R}^{md \times hw}$& Standardizing   &  Learnable $\gamma,\beta \in \R^{d}$  & Arxiv, 2016   \\
	\hline
$L^p$-Norm BN~\cite{2016_arxiv_Liao}   & $\Pro_{BN}(\tX{} )\in \mathbb{R}^{d \times mhw} $& Standardizing with $L^p$-Norm divided  &  Learnable $\gamma,\beta \in \R^{d}$ & Arxiv, 2016   \\
	\hline
Divisive normalization~\cite{2017_ICLR_Ren}   & $\Pro_{LN}(\tX{} ) \in \mathbb{R}^{m \times dhw}$: local spatial positions and channels& Standardizing   &   Learnable $\gamma,\beta \in \R^{d}$  & ICLR, 2017   \\
		\hline
Conditional IN~\cite{2017_ICLR_Dumoulin}   & $\Pro_{IN}(\tX{} ) \in \mathbb{R}^{md \times hw}$& Standardizing   &  Side information & ICLR, 2017   \\
	\hline	
Dynamic LN~\cite{2017_InterSpeech_Kim}  & $\Pro_{LN}(\tX{} )\in \mathbb{R}^{m \times dhw} $& Standardizing   &  Generated $\gamma,\beta \in \R^{d}$ & INTERSPEECH, 2017   \\
	\hline
Conditional BN~\cite{2017_NeurIPS_Vries}   & $\Pro_{BN}(\tX{} )\in \mathbb{R}^{d \times mhw} $& Standardizing   &  Side information & NeurIPS, 2017   \\
		\hline
Pixel normalization~\cite{2018_ICLR_Karras}  & $\Pro_{PN}(\tX{} )\in \mathbb{R}^{mhw \times d }$& Scaling   &  No & ICLR, 2018   \\
	 \hline
Decorrelated BN~\cite{2018_CVPR_Huang}   & $\Pro_{BN}(\tX{} )\in \mathbb{R}^{d \times mhw} $& ZCA whitening   &   Learnable $\gamma,\beta \in \R^{d}$  & CVPR, 2018   \\
	 \hline
Group normalization (GN)~\cite{2018_ECCV_Wu}   & $\Pro_{GN}(\tX{} ) \in \mathbb{R}^{mg \times shw}$& Standardizing   &   Learnable $\gamma,\beta \in \R^{d}$  & ECCV, 2018   \\
	\hline
Adaptive IN~\cite{2018_ECCV_Huang}   & $\Pro_{IN}(\tX{} ) \in \mathbb{R}^{md \times hw}$& Standardizing   &   Generated $\gamma,\beta \in \R^{d}$ & ECCV, 2018   \\
	\hline
$L^1$-Norm BN~\cite{2018_NeurIPS_Hoffer,2018_arxiv_Wu}  & $\Pro_{BN}(\tX{} )\in \mathbb{R}^{d \times mhw} $& Standardizing with $L^1$-Norm divided  &   Learnable $\gamma,\beta \in \R^{d}$  & NeurIPS, 2018   \\
\hline
Whitening and coloring BN~\cite{2019_ICLR_Siarohin}   & $\Pro_{BN}(\tX{} ) \in \mathbb{R}^{d \times mhw}$& CD whitening   &  Color transformation & ICLR, 2019   \\
\hline
Generalized BN~\cite{2019_AAAI_Yuan}  & $\Pro_{BN}(\tX{} )\in \mathbb{R}^{d \times mhw} $& General standardizing   &   Learnable $\gamma,\beta \in \R^{d}$  & AAAI, 2019   \\
	\hline	
		Iterative normalization~\cite{2019_CVPR_Huang}   & $\Pro_{BN}(\tX{} )\in \mathbb{R}^{d \times mhw} $& ZCA whitening by Newton's
		iteration   &   Learnable $\gamma,\beta \in \R^{d}$  & CVPR, 2019   \\
		\hline
Instance-level meta normalization~\cite{2019_CVPR_Jia}   & $\Pro_{LN}(\tX{} )/\Pro_{IN}(\tX{} )/$ $\Pro_{GN}(\tX{} ) $& Standardizing   &  Learnable $\&$ generated $\gamma,\beta \in \R^{d}$ & CVPR, 2019   \\
	\hline
Spatially adaptive denormalization~\cite{2019_CVPR_Park}   & $\Pro_{BN}(\tX{} ) \in \mathbb{R}^{d \times mhw}$& Standardizing   &  Generated $\gamma,\beta \in \R^{d \times h \times w}$ & CVPR, 2019   \\
		\hline
Position normalization (PN) ~\cite{2019_NeurIPS_Li}  & $\Pro_{PN}(\tX{} ) \in \mathbb{R}^{mhw \times d}$& Standardizing   &   Learnable $\gamma,\beta \in \R^{d}$  & NeurIPS, 2019   \\
	\hline
	Root mean square LN~\cite{2019_NeurIPS_Zhang}  & $\Pro_{LN}(\tX{} ) \in \mathbb{R}^{m \times dhw}$& Scaling   &   Learnable $\gamma \in \R^{d}$  & NeurIPS, 2019   \\
	\hline
	Online normalization~\cite{2019_NeurIPS_Chiley}  & $\Pro_{LN}(\tX{} )\in \mathbb{R}^{m \times dhw} $& Scaling   &  No & NeurIPS, 2019   \\
	\hline
Batch group normalization~\cite{2020_ICLR_Summers}  & $\Pro_{BGN}(\tX{} ) \in \mathbb{R}^{g_m g \times s_m shw}$& Standardizing   &   Learnable $\gamma,\beta \in \R^{d}$  & ICLR, 2020   \\
	\hline
Instance enhancement BN~\cite{2020_AAAI_Liang}  & $\Pro_{BN}(\tX{} ) \in \mathbb{R}^{d \times mhw}$& Standardizing   &  Learnable $\&$ generated $\gamma,\beta \in \R^{d}$ & AAAI, 2020   \\
	\hline	
PowerNorm~\cite{2020_ICML_Shen}  & $\Pro_{BN}(\tX{} ) \in \mathbb{R}^{d \times mhw}$& Scaling   &  Learnable  $\gamma,\beta \in \R^{d}$ & ICML, 2020   \\
	\hline	
Local contex normalization~\cite{2019_arxiv_Ortiz} & $\Pro_{GN}(\tX{} )\in \mathbb{R}^{mg \times shw}$: local spatial positions and channels & Standardizing   &   Learnable $\gamma,\beta \in \R^{d}$  & CVPR, 2020   \\
\hline
	Filter response normalization~\cite{2020_CVPR_Singh}   & $\Pro_{IN}(\tX{} ) \in \mathbb{R}^{md \times hw}$& Scaling   &   Learnable $\gamma,\beta \in \R^{d}$  & CVPR, 2020   \\
	\hline
Attentive normalization~\cite{2019_arxiv_Li}  & $\Pro_{BN}(\tX{} )/\Pro_{IN}(\tX{} )/$ $\Pro_{LN}(\tX{} ) /\Pro_{GN}(\tX{} )  $& Standardizing   &   Generated $\gamma,\beta \in \R^{d}$ & ECCV, 2020   \\
	\hline
			\toprule[1pt]
		\end{tabular}
	\end{small}
	\vspace{-0.20in}
\end{table*}

\subsection{Multi-Mode and Combinational Normalization}
In previous sections, we focused on single-mode normalization methods. In this section, we will introduce the methods that extend to multiple modes, as well as combinational methods.

\vspace{0.05in}
\Para{Multiple Modes:}
%
Kalayeh and Shah proposed mixture normalizing (MixNorm)  ~\cite{2019_TPAMI_Kalayeh}, which performs normalization on subregions that can be identified by disentangling the different modes of the distribution, estimated via a Gaussian mixture
model (GMM).  MixNorm requires a two-stage process, where the GMM is first fitted by  expectation-maximization (EM)~\cite{1977_JS_Dempster} with K-means++~\cite{2007_ASDA_Arthur}  for initialization, and the normalization is then performed on samples with respect to the estimated parameters. MixNorm is not fully differentiable due to the K-means++ and EM iterations.

Deecke \etal proposed mode normalization (ModeNorm), which also extends the normalization to more than one  mean and variance to address the heterogeneous nature of complex datasets. MN is formulated in a mixture of experts (MoE) framework, where a set of simple gate functions is introduced to assign one example to groups with a given probability. Each sample in the mini-batch is then normalized under voting from its gate assignment. The gate functions are trained jointly by backpropagation. 

%
%

\vspace{0.05in}
\Para{Combination:}
Since different normalization strategies have different advantages and disadvantages for training DNNs, some methods try to combine them.
Luo \etal proposed switchable normalization (SN)~\cite{2019_ICLR_Luo}, which combines  three types of statistics, estimated channel-wise, layer-wise, and mini-batch-wise, by using IN, LN, and BN, respectively. SN switches between the different normalization methods by learning their importance weights, computed by a softmax function. SN was designed to address the learning-to-normalize problem and obtains good results on several visual benchmarks~\cite{2019_ICLR_Luo}.
 Shao \etal~\cite{2019_CVPR_Shao} further introduced sparse switchable normalization (SSN), which selects different normalizations using the proposed SparsestMax function, which is a sparse version of softmax.
  Pan \etal~\cite{2019_ICCV_Pan} proposed switchable whitening (SW), which provides a general way to switch between different whitening and standardization methods under the SN framework.
  Zhang \etal~\cite{2020_CVPR_Zhang} introduced exemplar normalization (EN) to  investigate a dynamic `learning-to-normalize' problem. EN learns different data-dependent normalizations for different image samples, while SN fixes the importance ratios for the entire dataset.
Besides, Luo \etal~\cite{2019_ICML_Luo} proposed dynamic normalization (DN), which generalizes IN, LN, GN and BN in a unified formulation and can interpolate them to produce new normalization methods.

Considering that IN can learn style-invariant features~\cite{2016_arxiv_Ulyanov}, Nam \etal~\cite{2018_NeurIPS_Nam} introduced batch-instance normalization (BIN) to normalize the styles adaptively to the task and selectively to individual feature maps. It learns to control how much of the style information is propagated through each channel by leveraging a learnable gate parameter to balance between IN and BN. A similar idea was also used in the adaptive layer-instance normalization (AdaLIN)~\cite{2020_ICLR_Kim} for image-to-image translation tasks, where a learnable gate parameter is leveraged to balance between LN and IN. Bronskill \etal ~\cite{2020_ICML_Bronskill} introduced TaskNorm, which combines LN/IN with BN for meta-learning scenarios.
 Rather than designing a combinational normalization module, Pan \etal proposed IBN-Net, which carefully integrates IN and BN as building blocks, and  can be wrapped into several deep networks to improve their performances. Qiao \etal introduced batch-channel normalization (BCN), which integrates BN and channel-based normalizations (e.g., LN and GN) sequentially as a wrapped module.

Recently, Liu \etal~\cite{2020_arxiv_Liu}  searched for a combination of normalization-activation layers using AutoML~\cite{2017_ICLR_Baker}, leading to the discovery of EvoNorms, a set of new normalization-activation layers with sometimes surprising structures that go beyond existing design patterns.

%


%
%

\subsection{BN for More Robust Estimation}
As illustrated in previous sections, BN introduces inconsistent normalization operations during training (using mini-batch statistics, as shown in \Eqn~\ref{eqn-BN-train}) and inference (using population statistics estimated in \Eqn~\ref{eqn-BN-running_average}). This means that the upper layers are trained on representations different from those computed during inference. These differences become significant if the batch size is too small, since the estimates of the mean and variance become less accurate. This leads to significantly degenerated performance~\cite{2017_NeurIPS_Ioffe,2018_ECCV_Wu,2019_ICCV_Singh,2020_arxiv_Kaku}. To address this problem, some normalization methods avoid normalizing along the batch dimension, as introduced in previous sections. Here, we will discuss the  more robust estimation methods that also address this problem of BN.
\subsubsection{Normalization as Functions Combining Population Statistics}

One way to reduce the discrepancy between training and inference is to combine the estimated population statistics for normalization during training.

Ioffe \etal~\cite{2017_NeurIPS_Ioffe} proposed batch renormalization (BReNorm), augmenting the normalized output for each neuron with an affine transform, as:
\begin{equation}
\label{eqn-BNReNorm}
\hat{x}=\frac{x - \mu}{\sigma} \cdot r + z,
\end{equation}
where $r=\frac{\sigma}{\hat{\sigma}}$ and $z=\frac{\mu - \hat{\mu}}{\hat{\sigma}}$. Note that $r$ and $z$ are bounded between $(\frac{1}{r_{max}}, r_{max})$ and $(\frac{1}{z_{max}}, z_{max})$, respectively. Besides, $r$ and $z$ are treated as constants when performing gradient computation. \Eqn~\ref{eqn-BNReNorm} is reduced to standardizing the activation using the estimated population (which ensures that the training and inference are consistent)  if $r$ and $z$ are between their bounded values. Otherwise, \Eqn~\ref{eqn-BNReNorm}  implicitly exploits the benefits of mini-batch statistics.

Dinh \etal~\cite{2017_ICLR_Dinh} were the first to experiment with batch normalization  using population statistics, which were weighted averages of the old population statistics and current mini-batch statistics, as shown in \Eqn~\ref{eqn-BN-running_average}. The experimental results demonstrate that,  combining population  and mini-batch statistics can improve the performance of BN in  small-batch-size scenarios. This idea is also used in diminishing batch normalization~\cite{2017_arxiv_Ma}, full normalization (FN)~\cite{2019_AISTATS_Lian}, online normalization~\cite{2019_NeurIPS_Chiley }, moving average batch normalization (MABN)~\cite{2020_ICLR_Yan},  PowerNorm~\cite{2020_ICML_Shen} and momentum batch normalization (MBN)~\cite{2020_ECCV_Yong2}. One challenge for this type of method is how to calculate the gradients during backpropagation, since the population statistics are computed by all the previous mini-batches, and it is impossible to obtain their exact  gradients~\cite{2016_arxiv_Liao}. One straightforward strategy is to view the population statistics as constant and only back-propagate through current mini-batches, as proposed in~\cite{2017_ICLR_Dinh},~\cite{2017_arxiv_Ma} and~\cite{2019_AISTATS_Lian}. However, this  may introduce  training instability, as discussed in Section~\ref{sec:normalizing-activation-population}. Chiley \etal ~\cite{2019_NeurIPS_Chiley }  proposed to compute the gradients by maintaining the property of BN during backpropagation. Yan \etal \cite{2020_ICLR_Yan} and Shen \etal \cite{2020_ICML_Shen} proposed to view the backpropagation gradients as statistics to be estimated, and approximate these statistics by moving averages.




Rather than explicitly using the population statistics,
Guo \etal ~\cite{2018_AAAI_Guo}  introduced  memorized BN, which considers data information from multiple recent batches (or all batches in an extreme case) to produce more accurate and stable statistics. 
A similar idea is exploited in cross-iteration batch normalization~\cite{2020_arxiv_Yao}, where the mean and variance of examples from recent iterations are approximated for the current network weights via a low-order Taylor polynomial.
Besides, Wang \etal proposed Kalman normalization~\cite{2018_NeurIPS_Wang}, which  treats all the layers in a network as a whole system, and estimates the statistics of a certain layer by considering the distributions of all its
preceding layers, mimicking the merits of Kalman filtering.
Another practical approach for relieving the small-batch-size issue of BN in engineering systems is the synchronized batch normalization~\cite{2017_CVPR_Zhao,2018_CVPR_Liu,2018_CVPR_Peng}, which performs a synchronized computation of BN statistics across GPUs (Cross-GPU BN) to obtain better statistics.

\subsubsection{Robust Inference Methods for BN}
Some works address the small-batch-size problem of BN by finely estimating corrected normalization statistics during inference only. This strategy does not affect the training scheme of the model.

In fact, even the original BN paper~\cite{2015_ICML_Ioffe} recommended estimating the population statistics after the training has finished (Algorithm 2 in \cite{2015_ICML_Ioffe}), rather than using the
estimation calculated by running average, as shown in \Eqn~\ref{eqn-BN-running_average}.
However, while this can benefit a model trained with a small batch size, where estimation is the main issue\cite{2015_ICML_Ioffe,2018_UAI_Izmailov,2019_ICLR_Luo}, it may lead to degenerated  generalization when the batch size is moderate. 

Singh \etal analyzed how  a small batch size hampers the estimation accuracies of BN when using running averages, and proposed EvalNorm~\cite{2019_ICCV_Singh}, which optimizes the sample weight  during inference to ensure that the activations produced by normalization are similar to those provided during training. A similar idea is also exploited in
\cite{2020_ICLR_Summers}, where the sample weights are viewed as hyperparameters, which are optimized on a validation set.

Compared to estimating the BN statistics (population mean and standardization deviation), Huang \etal ~\cite{2020_CVPR_Huang} showed that estimating the whitening matrix of BW is more challenging. They demonstrated that, in terms of estimating the population statistics of the whitening matrix, it is more stable to use the mini-batch covariance matrix indirectly (the whitening matrix can be calculated after training) than the mini-batch whitening matrix directly.


\section{Normalizing Weights}
\label{sec:normalizing-weight}
As stated in Section~\ref{sec:motivaiton-overview}, normalizing the weights can implicitly normalize the activations by imposing  constraints on the weight matrix, which can contribute to preserving the activations (gradients) during forward (backpropagation).
 Several seminal works have analyzed the distributions of the activations, given normalized inputs, under the assumption that the weights have certain properties or are under certain constraints,
\eg,  normalization propagation~\cite{2016_ICML_Arpit}, variance propagation~\cite{2018_arxiv_Shekhovtsov}, self normalization~\cite{2017_NeurIPS_Klambauer}, bidirectional self-normalization~\cite{2020_arxiv_Lu}. The general idea of weight normalization is to provide  layer-wise constraints on the weights during optimization, which can be formulated as:
  \begin{eqnarray}
\label{eqn:optim-constraints}
	\theta^* &=\arg \min_{\theta} \mathbb{E}_{(\rvx,\rvy)\in D} [\mathcal{L}(\rvy, f(\rvx;\theta))] \nonumber \\
   & s.t.~~~~~ \Upsilon (\mW),
\end{eqnarray}
where $\Upsilon (\mW)$ are the layer-wise constraints imposed on the weights $\mW \in \R^{d_{out} \times d_{in}}$. It has been shown that the imposed constraints can benefit generalization~\cite{2017_ICCV_Huang,2018_AAAI_Huang,2018_AAAI_Mete}. In the following sections, we will introduce different constraints and discuss how to train a model with the constraints satisfied.

\subsection{Constraints on Weights}
Salimans \etal proposed weight normalization (WN), which requires the input weight of each neuron to be unit norm. Specifically, given one neuron's input weight $\mW_{i} \in \R^{d_{in}}$, the constraints imposed on  $\mW$ are:
  \begin{eqnarray}
  \label{eqn:constraints-WN}
\Upsilon (\mW)= \{\| \mW_{i} \|=1, i=1,...,d_{out} \}.
  \end{eqnarray}
Weight normalization has a scale-invariant property like BN, which is important  for stabilizing training.

Inspired by the practical weight initialization technique~\cite{2010_AISTATS_Glorot,2015_ICCV_He}, where  weights are sampled
from a distribution with zero mean and a standard deviation
for initialization, Huang \etal~\cite{2017_ICCV_Huang} further proposed centered weight normalization (CWN), constraining the input weight of each neuron to have zero mean and unit norm, as:
  \begin{eqnarray}
  \label{eqn:constraints-CWN}
\Upsilon (\mW)= \{\mW_{i}^T \mathbf{1}=0~\&~\|    \mW_{i} \|=1, i=1,...,d_{out} \}.
  \end{eqnarray}
  CWN can theoretically preserve the activation statistics between different layers under certain assumptions (Proposition 1 in ~\cite{2017_ICCV_Huang}), which can benefit optimization. Weight centering is also advocated for in~\cite{2019_arxiv_Qiao,2020_ICLR_Yan}. Qiao \etal proposed weight standardization (WS), which imposes the constraints on the weights~\cite{2020_AAAI_Li} with   $\Upsilon (\mW)= \{\mW_{i}^T \mathbf{1}=0~\&~\|    \mW_{i} \|=\sqrt{d_{out}}, i=1,...,d_{out} \}$. Note that WS cannot effectively  preserve the activation statistics between different layers, since the weight norm is $\sqrt{d_{out}}$, which may cause exploding activations. Therefore, WS usually needs to be combined with activation normalization methods (\eg, BN/GN) to  relieve this issue~\cite{2019_arxiv_Qiao}.

Another widely used constraint on weights is orthogonality, which is represented as
  \begin{eqnarray}
  \label{eqn:constraints-OWN}
  \Upsilon (\mW)= \{\mW \mW^T= \mI \}.
  \end{eqnarray}
Orthogonality was first used in the square hidden-to-hidden weight matrices  of RNNs~\cite{2016_ICML_Arjovsky,2016_NeurIPS_Wisdom,2016_CoRR_Dorobantu,2017_ICML_Eugene,2017_AAAI_Hyland,2017_GRU_Jing,2018_ICML_Kyle}, and then further extended to the more general rectangular matrices in DNNs~\cite{2018_AAAI_Huang,2019_NeurIPS_Mete,2019_TPAMI_Jia, 2020_CVPR_WangJiayun,2020_ICML_Qi}. Orthogonal weight matrices can theoretically preserve the norm of activations/output-gradients between linear transformations~\cite{2020_CVPR_Huang2,2020_CVPR_WangJiayun,2020_ICML_Qi}. Further, the distributions of activations/output-gradients can also be preserved under mild assumptions~\cite{2018_AAAI_Huang,2020_CVPR_Huang2}. These properties of orthogonal weight matrices are beneficial for the optimization of DNNs. Furthermore, orthogonal weight matrices can avoid learning redundant filters, benefitting generalization.

Rather than bounding all singular values as 1, like in orthogonal weight matrices, Miyato \etal~\cite{2018_ICLR_Miyato} proposed spectral normalization, which constrains the spectral norm (the maximum singular value) of a weight matrix to 1,  in order to control the Lipschitz constant of the discriminator when training GANs.
Huang \etal~\cite{2020_CVPR_Huang2} proposed orthogonalization by Newton's iterations (ONI), which controls the orthogonality through the iteration number. They showed that it is possible to bound the singular values of a weight matrix  between $(\sigma_{min}, 1)$ during training.  ONI effectively interpolates between spectral normalization and full orthogonalization, by altering the iteration number.

Note that the  constraints imposed on the weight matrix (Eqns.~\ref{eqn:constraints-WN}, \ref{eqn:constraints-CWN}, \ref{eqn:constraints-OWN}) may harm the representation capacity and result in degenerated performance. An extra learnable scalar parameter is usually used for each neuron to recover the possible loss in  representation capacity, which is similar to the idea of the affine parameters proposed in BN.


\subsection{Training with Constraints}
It is clear that training a DNN with constraints imposed on the weights is a constraint optimization problem. Here, we summarize three kinds of strategies for solving this.


\vspace{0.05in}
\Para{Re-Parameterization}
One stable way to solve constraint optimization problems is to use a re-parameterization method. Re-parameterization constructs a fine transformation $\psi$ over the proxy parameter $\mV$ to ensure that the transformed weight $\mW$ has certain beneficial properties for the training of neural networks. Gradient updating is executed on the proxy parameter $\mV$ by back-propagating the gradient information through the normalization process. Re-parameterization was first used in learning the square orthogonal weight matrices in RNNs~\cite{2016_ICML_Arjovsky,2016_NeurIPS_Wisdom,2016_CoRR_Dorobantu}. Salimans \etal~\cite{2016_NeurIPS_Salimans} used this  technique to learn a unit-norm constraint as shown in Eqn.~\ref{eqn:constraints-WN}. Huang \etal~\cite{2017_ICCV_Huang} formally described the re-parameterization idea in training with constraints on weight matrices, and applied it to solve the optimization with zero-mean and unit-norm constraints as shown in Eqn.~\ref{eqn:constraints-CWN}. This technique was then further used in other methods for learning with different constraints, \eg, orthogonal weight normalization~\cite{2018_AAAI_Huang}, weight standardization~\cite{2019_arxiv_Qiao}, spectral normalization~\cite{2018_ICLR_Miyato} and weight centralization~\cite{2020_ICLR_Yan}.
Re-parameterization is a main technique for optimizing  constrained weights in DNNs. Its main merit is that the training is relatively stable, because it updates  $\mV$ based on the gradients computed by backpropagation, while simultaneously maintaining the constraints on $\mW$. The downside is that the backpropagation through the designed transformation may increase the computational cost.
%
%
%
%
%

\vspace{0.05in}
\Para{Regularization with an Extra Penalty}
Some works have tried to maintain the weight constraints using an additional penalty on the objective function, which can be viewed as a regularization. This regularization technique is mainly used for learning the weight matrices with orthogonality constraints, for its efficiency in computation~\cite{2013_ICML_Pascanu,2017_ICML_Eugene,2017_CVPR_Xie,2018_NeurIPS_Bansal,2019_Arxiv_Amjad}.
Orthogonal regularization  methods have demonstrated improved performance  in image classification \cite{2017_CVPR_Xie,2018_NeurIPS_Zhang,2018_CVPR_Lezama,2018_NeurIPS_Bansal},  resisting  attacks from adversarial examples \cite{2017_ICML_Moustpha}, neural photo editing \cite{2017_ICLR_Brock} and training GANs \cite{2019_ICLR_Brock, 2018_ICLR_Miyato}.
However, the introduced penalty  works like a pure regularization, and whether or not the constraints are truly maintained or  training benefited is unclear. Besides, orthogonal regularization usually requires to be combined with activation normalization, when applied on  large-scale architectures, since it cannot stabilize training.


%

\vspace{0.05in}
\Para{Riemannian Optimization}
A weight matrix $\mW$ with constraints can be viewed as an embedded submanifold~\cite{2017_NeurIPS_Cho,2018_AAAI_Huang,2017_arxiv_Huang,2020_ICLR_LiJun}. For example, a weight matrix with an orthogonality constraint (\Eqn \ref{eqn:constraints-OWN}) is a real Stiefel manifold~\cite{2017_NeurIPS_Cho,2018_AAAI_Huang}. One possible way of maintaining these constraints when training DNNs is to use Riemannian optimization~\cite{2016_Corr_Ozay,2017_Corr_Harandi}.
Conventional Riemannian  optimization techniques are based on a gradient descent method over a manifold, which iteratively seeks updated points. In each iteration, there are two main phases: 1) Computing the Riemannian gradient  based on the inner dot product defined in  the  tangent space of the manifold; 2) Finding the descent direction and ensuring that the new point is on the manifold~\cite{2017_Corr_Harandi,2018_AAAI_Huang}.
Most Riemannian optimization methods in the deep learning community address the second phase by either QR-decomposition-type retraction\cite{2013_TSP_Kaneko,2017_Corr_Harandi} or Cayley transformation~\cite{2013_MP_Wen,2016_NeurIPS_Wisdom,2017_ICML_Eugene,2020_ICLR_LiJun}.
The main difficulties of applying Riemannian optimization in training DNNs are: 1) The optimization space covers multiple embedded submanifolds; 2) The embedded submanifolds are inter-dependent since the optimization of the current weight layer is affected by those of preceding layers. To stabilize the training, activation normalizations (\eg BN)~\cite{2020_ICLR_LiJun} or gradient clips~\cite{2017_NeurIPS_Cho} are usually required. One interesting observation is that using BN will probably improve the performance of projection-based methods (where the gradient is calculated based on the Euclidean space)~\cite{2017_CVPR_Jia,2017_arxiv_Huang}.

\subsection{Combining Activation Normalization}
Normalizing weights has its own advantages in training DNNs, compared to normalizing activations. For example, it is data-independent, and is more convenient to use for theoretical analysis \cite{2015_COLT_Neyshabur,2018_arxiv_WuXiaoxia}. However, normalizing weights has some drawbacks when used in large-scale networks in practice: 1) It may not effectively improve the optimization efficiency when  residual connections are introduced or when the nonlinearity does not satisfy the assumptions required for preserving distributions between different layers, since the Criteria 1 in Section~\ref{sec:motivaiton-overview} is not readily satisfied in these situations; 2) Normalizing the weight usually has significantly lower test accuracy than BN on large-scale image classification~\cite{2017_arxiv_Gitman}.
Huang \etal showed that CWN combined with BN can improve the original networks with only BN. The idea of combining normalizing weights and activations to improve performance has been widely studied~\cite{2018_AAAI_Huang,2019_arxiv_Qiao,2019_arxiv_Qiao2,2020_CVPR_Huang2}. Moreover, Luo \etal proposed cosine normalization~\cite{2018_ICANN_Luo}, which merges layer normalization and weight normalization together.


\section{Normalizing Gradients}
\label{sec:normalizing-gradient}
As stated previously, normalizing activations and weights aims to provide a better optimization landscape for DNNs, by satisfying Criteria 1 and 2 in Section~\ref{sec:motivaiton-overview}. Rather than providing a good optimization landscape by design, normalizing gradients in DNNs aims to exploit the  curvature information for GD/SGD, even though the optimization landscape is ill-conditioned.  It performs normalization solely on the gradients, which may effectively remove the negative effects of an ill-conditioned landscape caused by the diversity in magnitude of gradients from different layers~\cite{2010_AISTATS_Glorot}. Generally speaking, normalizing gradients is  similar to second-order optimization~\cite{2010_ICML_Martens,2012_AISTATS_Vinyals,2012_NN_Martens,2015_ICML_Grosse} or coordinate-wise adaptive learning rate based methods~\cite{2011_JMLR_Duchi,2012_arxiv_Hinton,2014_arxiv_Kingma}, but with the goal of exploiting the layer-wise structural information in DNNs.

  Yu \etal~\cite{2017_arxiv_Yu} were the first to propose block-wise (layer-wise) gradient normalization for training DNNs to front the gradient explosion or vanishing problem. Specifically, they perform scaling over the gradients \wrt the weight  in each layer, ensuring the norm to be unit-norm. This technique can  decrease the magnitude of a large gradient to a certain level, like gradient clipping~\cite{2013_ICML_Pascanu}, and also boost the magnitude of a small gradient.
  However, the net-gain of this approach degenerates in the scale-invariant DNNs (\eg with BN). In~\cite{2017_arxiv_Yu}, an extra ratio factor that depends on the norm of the layer-wise weight was used to adaptively adjust the magnitude of the gradients.  A similar idea was also introduced in the layer-wise adaptive  rate scaling (LARS), proposed by You \etal~\cite{2017_arxiv_You}, for large-batch training. LARS and its follow-up works~\cite{2020_ICLR_You,2020_arxiv_Zheng,2020_arxiv_Huo} are  essential techniques in training large-scale DNNs using large batch sizes, significantly reducing the  training times without degradation of performance.

  Rather than using the scaling operation, Yong \etal ~\cite{2020_ECCV_Yong} recently proposed
  gradient centralization (GC), which performs centering over the gradient \wrt the input weight of each neuron in each layer. GC implicitly imposes constraints on the input weight, and ensures that the sum of elements in the input weight is a constant during training.  GC effectively improves the performances of DNNs with activation normalization (\eg BN or GN).

\section{Analysis of Normalization}
\label{sec:theory}
In Section~\ref{sec:motivaiton-overview}, we provided  high-level motivation of normalization in benefiting network optimization. In this section, we will further discuss other properties of normalization methods in improving DNNs' training performance. We mainly focus on BN, because it displays nearly all the benefits of normalization in improving the performance of DNNs, \eg, stabilizing training, accelerating convergence and improving the generalization.

\subsection{Scale Invariance in Stabilizing Training}
\label{sec:theory-scale-invariant}
One essential functionality of BN is its ability to stabilize  training. This  is mainly due to its scale-invariant property~\cite{2016_NeurIPSW_Ba,2016_ICLR_Neyshabur,2020_AAAI_Sun,2020_arxiv_Wan}, \ie, it does not change the prediction when rescaling parameters and works by adaptively adjusting the learning rate in a layer-wise manner~\cite{2017_NeurIPS_Cho,2018_NeurIPS_Hoffer,2019_ICLR_Arora}. Specifically,
\begin{eqnarray}
\label{eqn:scale-invarint}
BN(\rvx; a \mW)=BN(\rvx; \mW) \\
\frac{\partial BN(\rvx; a \mW) }{\partial (a \mW)}=\frac{1}{a}   \frac{\partial BN(\rvx; \mW) }{\partial  \mW},
\end{eqnarray}
 where $a$ is a constant factor.
 The scale-invariant property also applies to other methods that normalize the activations~\cite{2016_NeurIPSW_Ba,2018_ECCV_Wu,2018_CVPR_Huang,2016_arxiv_Ulyanov} or weights~\cite{2016_NeurIPS_Salimans,2017_ICCV_Huang,2018_AAAI_Huang}.
 This property was first shown in the original BN paper, and then investigated in~\cite{2016_NeurIPSW_Ba} to compare different normalization methods, and further extended for rectifier networks in~\cite{2020_ECCV_Huang}. Specifically, in~\cite{2020_ECCV_Huang}, Huang \etal showed how the scaled factor $a$ of the weight in a certain layer will lead to  exponentially increased/decreased gradients for unnormalized rectifier networks, and how normalization can avoid this problem with its scale-invariant property.

  The scale-invariant weight vector in a network is always perpendicular to its gradient~\cite{2016_ICLR_Neyshabur,2016_NeurIPS_Salimans,2019_ICLR_Arora,2020_AAAI_Sun,2020_arxiv_Roburin}, which has an auto-tuning effect~\cite{2018_arxiv_Wu,2019_ICLR_Arora,2019_ICML_Cai,2020_arxiv_Chai}.
 Wu \etal proposed WNgrad ~\cite{2018_arxiv_WuXiaoxia}, which
 introduces an adaptive step-size algorithm based on this fact.
 In ~\cite{2019_ICLR_Arora}, Arora \etal proved that GD/SGD with BN can arrive a first-order stationary point with any  fixed learning rate, under certain mild assumptions.
Cai \etal~\cite{2019_ICML_Cai} showed that, for the simple problem of ordinary least squares (OLS), GD with BN converges under arbitrary learning rates for the weights, and the convergence remains linear under mild conditions.

 Another  research direction is to analyze the effect of weight decay~\cite{1992_WD_Krogh} when combined with scale-invariant normalization methods~\cite{2017_arxiv_Laarhoven,2017_arxiv_Huang,2018_NeurIPS_Hoffer,2019_ICLR_Zhang,2020_AAAI_Li,2020_ICLR_Li,2020_arxiv_Wan}. In this case, weight decay causes the parameters to have smaller norms, and thus the effective learning rate is larger. In~\cite{2020_ICLR_Li}, Li and Arora showed that the original learning rate schedule and  weight decay can be folded into a new exponential schedule, when scale-invariant normalization methods are used.

%
%
%
%


\subsection{Improved Conditioning in Optimization}
\label{sec:theory-optimization}
As stated in previous sections, one motivation behind BN is that whitening the input can improve the conditioning of the optimization \cite{2015_ICML_Ioffe}  and thus accelerate training \cite{2015_NeurIPS_Desjardins,2018_CVPR_Huang}.
This motivation is theoretically supported for linear models~\cite{1990_NeurIPS_LeCun,2020_ICLR_Li}, but is difficult to further extend to DNNs.
In~\cite{2018_NeurIPS_Santurkar}, Santurkar \etal argue that BN may improve optimization by enhancing the smoothness of the  Hessian of the loss. However, this conclusion is based on a layer-wise analysis~\cite{2018_NeurIPS_Santurkar,2020_ECCV_Huang}, which corresponds to the diagonal blocks of the overall Hessian. Ghorbani \etal~\cite{2019_ICML_Ghorbani} further empirically investigated the conditioning of the optimization problem by computing the spectrum of the Hessian for a large-scale dataset. It is believed that the improved conditioning enables large learning rates for training, thus improving the generalization, as shown in \cite{2018_NeurIPS_Bjorck}. Karakida \etal~\cite{2019_NeurIPS_Karakida} investigated the conditioning of the optimization problem by analyzing the geometry of the parameter space determined by the Fisher information matrix (FIM), which also corresponds to the local shape of the loss landscape under certain conditions.

One intriguing phenomenon is that the theoretical benefits of whitening the input for optimization only hold when BN is placed before the linear layer, while, in practice, BN is typically placed after the linear layer, as recommended in \cite{2015_ICML_Ioffe}. In~\cite{2020_ECCV_Huang}, Huang \etal experimentally observed, through a layer-wise conditioning analysis, that BN (placed after the linear layer) not only improves the conditioning of the activation's covariance matrix, but also improves the conditioning of the output-gradient's covariation. Similar observations were made in~\cite{2020_arxiv_Daneshmand}, where BN prevents the rank collapse of pre-activation matrices. Some works have also empirically investigated the position, at which BN should be plugged in \cite{2016_ICLR_Mishkin,2019_arxiv_Chen,2018_CVPR_Huang}. Results have shown that placing it after the linear layer may work better, in certain situations.

Other analyses of normalization in optimization include an investigation into  the signal propagation and gradient backpropagation~\cite{2019_ICLR_Yang,2019_arxiv_Wei,2019_ICML_Labatie}, based on the mean field theory~\cite{2018_ICLR_Lee,2019_ICLR_Yang,2019_arxiv_Wei}.
Besides, the work of ~\cite{2019_AISTATS_Kohler} demonstrated that BN obtains an accelerated convergence on the (possibly) nonconvex problem of learning half-spaces with Gaussian
inputs, from a length-direction decoupling perspective. Dukler \etal~\cite{2020_ICML_Dukler} further provided the  first global convergence result for two-layer neural networks with ReLU~\cite{2010_ICML_Nair} activations trained with weight normalization.


%
%


\subsection{Stochasticity for Generalization}
One important property of BN is its ability to improve the generalization of DNNs. It is believed such an improvement is obtained from the stochasticity/noise introduced by normalization over batch data~\cite{2015_ICML_Ioffe,2018_ACCV_Alexander,2020_AAAI_Liang}.
It is clear that both the normalized output  (\Eqn \ref{eqn-BN-train}) and the population statistics  (\Eqn \ref{eqn-BN-running_average}) can be viewed as stochastic variables,  because they  depend on the mini-batch inputs, which are sampled over datasets. Therefore, the stochasticity comes from the normalized output during training~\cite{2019_CVPR_Huang}, and the discrepancy of normalization between training (using estimated population statistics)  and inference (using estimated population statistics)~\cite{2019_ICLR_Luo2,2020_CVPR_Huang}.

Ioffe and Szegedy~\cite{2015_ICML_Ioffe} were the first to show the advantages of this stochasticity for the generalization of networks, like dropout~\cite{2014_JMLR_Nitish,2019_CVPR_Li}. Teye \etal~\cite{2018_ICML_Teye} demonstrated that training a DNN using BN is equivalent to approximating inference in Bayesian models, and that uncertainty estimates can be obtained from any network using BN through Monte Carlo sampling during inference. This idea was further efficiently approximated by stochastic batch normalization~\cite{2018_ICLRW_Atanov} and exploited in prediction-time batch normalization~\cite{2020_arxiv_Nado}.
Alexander \etal jointly formulate the stochasticity of the normalized output and the discrepancy of normalization between training and inference in a mathematical way, under the assumptions that the distribution of  activations over the full dataset is approximately Gaussian and \textit{i.i.d}. In~\cite{2019_CVPR_Huang}, Huang \etal proposed an empirical evaluation for the stochasticity of normalization over batch data, called stochastic normalization disturbance (SND), and investigated how the batch size affects the stochasticity of BN. This  empirical analysis was further extended to the more general BW in~\cite{2020_CVPR_Huang}.

Some studies exploit the stochasticity of BN to improve the generalization for large-batch training, by altering the batch size when estimating the population statistics. One typical work is ghost batch normalization~\cite{2017_NeurIPS_Hoffer,2020_ICLR_Summers,2020_arxiv_Dimitriou}, which reduces the generalization error  by acquiring the statistics on small virtual (`ghost') batches instead of the real large batch.

%


%


\section{Applications of Normalization}
\label{sec:application}
As previously stated, normalization methods can be wrapped as general modules, which have been extensively integrated into various DNNs to stabilize and accelerate training, probably leading to improved generalization. For example, BN is an essential module in the state-of-the-art network architectures for CV tasks~\cite{2015_ImageNet,2016_CVPR_He,2016_BMVC_Zagoruyko,2016_CVPR_Szegedy,2017_CVPR_HuangGao,2017_CVPR_Xie}, and LN is an essential module in NLP tasks~\cite{2017_NeurIPS_Vaswani,2018_ICLR_Yu,2019_NeurIPS_Xu}.
In this section, we discuss the applications of normalization for particular tasks, in which normalization methods can effectively solve the key issues. To be specific, we mainly review the applications of normalization in domain adaptation, style transfer, training GANs and efficient deep models. However, we  note that there also exist works exploring how to apply normalization to meta learning~\cite{2018_arxiv_Nichol,2019_ICLR_Gordon,2020_ICML_John}, reinforcement learning~\cite{2016_NeurIPS_Hasselt,2019_arxiv_Bhatt,2020_ICML_Wang}, unsupervised representation learning~\cite{2020_CVPR_He,2020_CVPRW_Kocyigit}, permutation-equivariant networks~\cite{2018_CVPR_Yi,2020_CVPR_Sun}, graph neural networks~\cite{2020_arxiv_Cai}, ordinary differential equation (ODE) based networks~\cite{2020_arxiv_Gusak}, symmetric positive definite (SPD) neural networks~\cite{2019_NeurIPS_Brooks}, and guarding against adversarial attacks~\cite{2019_arxiv_Galloway,2020_ICLRSUB_Awais,2020_ICLR_Xie}.

\subsection{Domain Adaptation}
\label{sec:application-domain}
Machine learning algorithms trained on some given data (source domain) usually perform poorly when tested on data acquired under different settings (target domain). This is explained in domain adaptation as resulting from a shift between the distributions of the source and target domains. Most methods for domain adaptation thus aim to bridge the gap between these distributions.
A typical way of achieving this is to align the distributions of the source and target domains based on the mini-batch/population statistics of BNs~\cite{2017_arxiv_Li}.

Li \etal~\cite{2017_arxiv_Li} proposed the first work to exploit BNs in domain adaptation, named adaptive batch normalization (AdaBN), where the BN statistics for the source domain are calculated during training, and then those for the target domain are modulated during testing. AdaBN enables  domain-invariant features to be learnt without requiring additional loss terms and the extra associated parameters.
The hypothesis behind AdaBN is that the domain-invariant information is stored in the weight matrix of each layer, while the domain-specific information is represented by the statistics of the BN layer. However, this hypothesis may not always hold because the target domain is not exploited at the training stage. As a result, it is difficult to ensure that the statistics of the BN layers in the source and target domains correspond to their domain-specific information.

One way to overcome this limitation is to couple the network parameters for both target and source samples in the training stage, which has been the main research focus of several follow-up works inspired by AdaBN.
In~\cite{2017_ICCV_Carlucci}, Carlucci \etal proposed automatic domain alignment layers (AutoDIAL), which are embedded in different levels of the deep architecture to align the learned source and target feature distributions to a canonical one.
AutoDIAL exploits the source and target features during the training stage, in which an extra parameter is involved in each BN layer as the trade-off between the source and target domains.
Chang \etal further proposed domain-specific batch normalization (DSBN)~\cite{2019_CVPR_Chang}, where multiple branches of BN are used, each of which is exclusively in charge of a single domain.
DSBN learns domain-specific properties using only the estimated population statistics of BN and  learns domain-invariant representations with the other parameters in the network.
This method effectively separates domain-specific information for unsupervised domain adaptation.
Similar ideas have also been exploited in unsupervised adversarial domain adaptation in the context of semantic scene segmentation~\cite{2019_WACV_Romijnders} and adversarial examples for improving image recognition~\cite{2020_CVPR_Xie}.
In ~\cite{2019_CVPR_Roy}, Roy \etal further generalized DSBN by a domain-specific whitening transform (DWT), where the source and target data distributions are aligned using their covariance matrices. Wang \etal ~\cite{2019_NeurIPS_Wang} proposed transferable normalization (TransNorm), which also calculates the statistics of inputs from the source and target domains separately, while computing the channel transferability simultaneously. The normalized features then go through channel-adaptive mechanisms to re-weight the channels according to their transferability.

Besides  population statistics, Seo \etal also exploited the affine transform (\Eqn \ref{eqn-BN-recover-matrix}) of BN to represent the domain-specific information in their proposed
domain-specific optimized normalization (DSON)~\cite{2019_arxiv_Seo}. Moreover, DSON normalizes the activations by a weighted average of multiple normalization statistics (typically BN and IN), and keeps track of the normalization statistics of each normalization type if necessary, for each domain. DSON targets  domain generalization, where examples in the target domain cannot be accessed during training. This task is considered to be more challenging than unsupervised domain adaptation.

\subsubsection{Learning Universal Representations}
The idea of applying BN in domain adaptation can be further extended to the learning of universal representations~\cite{2017_arxiv_Bilen}, by constructing neural networks that work simultaneously in many domains. To achieve this, the networks need to learn to share common visual structures where no obvious commonality exists.  Universal representations cannot only benefit domain adaptation but also contribute to
multi-task learning, which aims to learn multiple tasks simultaneously in the same data domain.

Bilen \etal~\cite{2017_arxiv_Bilen} advocate to learn universal image representations using 1) the convolutional kernels to extract domain-agnostic information and 2) the BN layers to transform the internal representations to the relevant target domains.
Data \etal \cite{2018_ECCV_Data} exploited BN layers to learn discriminate visual classes, while other layers (\eg convolutional layers) are used to learn the universal representation. They also provided a way to interpolate BN layers to solve new tasks. Li \etal ~\cite{2019_CVPR_LiYunSheng}  proposed covariance normalization (CovNorm) for multi-domain learning, which provides efficient solutions to several tasks defined in different domains.

\subsection{Style Transfer}
\label{sec:application-style}
Style transfer is an important image editing task that enables the creation of new artistic works~\cite{2017_NeurIPS_Li,2019_TOCG_Jing}. Image style transform algorithms aim to generate a stylized image that has similar content and style to the given images. The key challenge in this task is to extract effective representations that can disentangle the style from the content. The seminal work by Gatys \etal \cite{2016_CVPR_Gatys} showed that the covariance/Gram matrix of the layer activations, extracted by a trained DNN, has a remarkable capacity for capturing visual styles.
This provides a feasible solution to matching the styles between images by minimizing Gram matrix based losses, pioneering the way for style transfer.

A key advantage of applying normalization to style transfer is that the normalization operation (NOP) can remove the style information (\eg, whitening can ensure the covariance matrix to be an identity matrix), while the normalization representation recovery (NRR), in contrast, introduces it. In other words, the style information is intuitively `editable' by normalization~\cite{2019_NeurIPS_Li,2020_arxiv_Li}. In a seminal work, Ulyanov \etal proposed instance normalization (IN)~\cite{2016_arxiv_Ulyanov} to remove instance-specific contrast information (style) from the content image. Since then, IN has been a basic module for image style transfer tasks.

In~\cite{2017_ICLR_Dumoulin}, Dumoulin \etal proposed conditional instance normalization (CIN), an efficient solution to integrating multiple styles. Specifically, multiple distinct styles are captured by a single network, by encoding the style information in the affine parameters (\Eqn \ref{eqn-BN-recover-matrix}) of IN layers, after which each style can be selectively applied to a target image. Huang \etal~\cite{2017_ICCV_HuangXun} proposed adaptive instance normalization (AdaIN), where the activations of content images are standardized by their statistics, and the affine parameters ($\beta$ and $\gamma$) come from the statistics of style activations. AdaIN transfers the channel-wise mean and variance feature statistics between content and style feature activations. AdaIN can also work well in text effect transfer, which aims at learning the visual effects while maintaining the text content~\cite{2020_AAAI_Liang}.
Rather than manually defining how to compute
the affine parameters so as to align the mean and variance between content and style features, dynamic instance normalization (DIN)~\cite{2020_AAAI_Jing}, introduced by Jing \etal, deals with arbitrary style transfer by encoding a style image into learnable convolution parameters, upon which the content image is stylized.

To address the limitations of AdaIN in only trying to match up the variances of the stylized image and the style image feature, Li \etal ~\cite{2017_NeurIPS_Li} further proposed whitening and coloring transformations (WCT)  to match up the covariance matrix. This shares a similar spirit to the optimization of the Gram matrix based cost in neural style transfer\cite{2017_NeurIPS_Li,2019_ICCV_Chiu}. Some methods ~\cite{2018_CVPR_Lu}  also seek to provide a good trade-off between AdaIN (which enjoys higher computational efficiency) and WCT (which synthesizes images visually closer to a given style).

\subsubsection{Image Translation}
In computer vision, image translation can be viewed as a more general case of image style transfer. Given an image in the source domain, the aim is to learn the conditional distribution of the corresponding images in the target domain. This includes, but is not limited to, the following tasks: super-resolution, colorization, inpainting, and attribute transfer. Similar to style transfer, AdaIN  is also an essential tool for image translation used in, for example, multimodal unsupervised image-to-image translation (MUIT)~\cite{2018_ECCV_Huang}. Note that the affine parameters of AdaIN in~\cite{2018_ECCV_Huang}  are produced by a learned network, instead of computed from statistics of a pretrained network as in ~\cite{2017_ICCV_HuangXun}.
Apart from IN, Cho \etal~\cite{2019_CVPR_Cho} proposed the group-wise deep whitening-and-coloring transformation (GDWCT)  by matching higher-order statistics, such as covariance, for image-to-image translation tasks. Moreover, since the whitening/coloring transformation can be considered a $1 \times 1$ convolution, Cho \etal \cite{2019_arxiv_Cho} further proposed adaptive convolution-based normalization (AdaCoN) to inject the target style into a given image, for unsupervised image-to-image translation. AdaCoN first performs standardization locally on each subregion of an input activation map (similar to the local normalization shown in Section~\ref{sec:normalizing-function-area}) and then applies an adaptive convolution, where the convolution filter weights are dynamically estimated using the encoded style representation.
Besides, Yu \etal proposed region normalization (RN)~\cite{2020_AAAI_Yu} for image inpainting network training. RN divides spatial pixels into different regions according to the input mask and standardizes the activations in each region. Wang ~\cite{2020_CVPR_Wang} introduced attentive normalization (AN) for conditional image generation, which is an extension of instance normalization~\cite{2016_arxiv_Ulyanov}. AN divides the feature maps into different regions based on their semantics, and then separately normalizes and denormalizes the feature points in the same region.

\subsection{Training GANs}
 GANs~\cite{2014_NeurIPS_goodfellow} can be regarded as a general framework to produce a model distribution that mimics a given target distribution. A GAN consists of a generator, which  produces the model distribution, and a discriminator, which distinguishes the model distribution from the target.
From this perspective, the ultimate goal when training GANs shares a similar spirit to model training in the domain adaptation task. The main difference lies in that GANs try to reduce the distance between different distributions, while domain adaptation models attempt to close the gap between different domains.
Therefore, the techniques that apply BN to domain adaptation, as discussed in Section~\ref{sec:application-domain}, may work for GANs as well.
For example, combining samples form different domains in a batch for BN may harm the generalization in domain adaptation, and this also applies for the training of GANs~\cite{2015_arxiv_Radford,2016_NeurIPS_Salimans2}.

One persisting challenge in training GANs is the performance control of the discriminator and the learning pace control between the discriminator and generator~\cite{2018_ICLR_Miyato}. The density ratio estimated by the discriminator is often inaccurate and unstable during training, and the generator may fail to learn the structure of the target distribution. One way to remedy this issue is to impose  constraints on the discriminator~\cite{2017_arxiv_Arjovsky}. For instance, Xiang \etal~\cite{2017_arxiv_Xiang} leveraged weight normalization to effectively improve the training performance of  GANs. Miyato \etal~\cite{2018_ICLR_Miyato} proposed spectral normalization (SN), which enforces Lipschitz continuity on the discriminator  by normalizing its parameters with the spectral norm estimated by power iteration. Since then, SN has become an important technique in training GANs\cite{2018_ICLR_Miyato,2019_ICML_Kurach,2019_ICLR_Brock}. In~\cite{2019_ICML_Zhang}, Zhang \etal further found that employing SN in the generator improves the stability, allowing for fewer training steps for the discriminator per iteration. Another important constraint in training GANs is the orthogonality~\cite{2019_ICLR_Brock,2020_CVPR_Huang2,2020_AAAI_Liu,2019_arxiv_Muller}. Brock \etal~\cite{2019_ICLR_Brock} found that applying orthogonal regularization to the generator renders it amenable to a simple `truncation trick', allowing fine control over the trade-off between sample fidelity and variety by reducing the variance of the generator input. Huang \etal~\cite{2020_CVPR_Huang2} proposed orthogonalization by Newton's iteration, which can effectively control the orthogonality of the weight matrix, and interpolate between spectral normalization and full orthogonalization by altering the iteration number.

As discussed in Section \ref{sec:application-style} for style transfer, the NRR operation of activation normalization can also be used as the side information for GANs, under the scenario of conditional GANs (cGANs)~\cite{2014_arxiv_Mirza}. cGANs  have shown  advancements in class conditional image generation~\cite{2017_ICML_Odena}, image generation from text~\cite{2016_ICML_Reed,2017_ICCV_Zhang}, and image-to-image translation~\cite{2017_ICCV_Zhu}.

In~\cite{2017_NeurIPS_Vries}, Vries \etal proposed conditional batch normalization (CBN), which injects a linguistic
input (e.g., a question in a VQA task) into the affine parameters of BN. This shares a similar spirit to the conditional instance normalization for style transfer,
and has been extensively explored in ~\cite{2018_ICLR_Miyato2,2019_ICML_Zhang,2019_ICLR_Brock,2019_arxiv_Michalski}.
In \cite{2019_CVPR_Karras}, Karras \etal proposed a style-based generator architecture for GANs, where the style information is embedded into the affine parameters of AdaIN~\cite{2017_ICCV_HuangXun}. Note that the style comes from the latent vector instead of an example image, enabling the model to work without external information.
Similarly, Chen \etal~\cite{2019_ICLR_Chen} proposed a more general self-modulation based on CBN, where the affine parameters can also be generated by the generator’s own input or provided by external information.

\subsection{Efficient Deep Models}
In real-world applications, it is essential to consider the efficiency of an algorithm in addition to its effectiveness due to the often limited computational resources (such as in smartphones). As such, there is also an active line of research exploiting normalization techniques (\eg, BN) to develop efficient DNNs based on network slimming or quantization. In network slimming, the general idea is to exploit the channel-wise scale parameter $\gamma \in \R^{d}$ of BN, considering that each scale $\gamma_i$ corresponds to a specific convolutional channel (or a neuron in a fully connected layer)~\cite{2017_ICCV_Liu}.
For example, Liu \etal~\cite{2017_ICCV_Liu} proposed to identify and prune insignificant channels (or neurons) based on the scale parameter in BN layers, which are imposed by $L^1$ regularization for sparsity. Ye \etal ~\cite{2018_ICLR_Ye} also adopted a similar idea, and developed a new algorithmic approach and rescaling trick to improve the robustness and speed of optimization. In~\cite{2020_ECCV_Li}, Li \etal proposed  an efficient evaluation component based on adaptive batch normalization~\cite{2017_arxiv_Li}, which has a strong correlation between different pruned DNN structures and their final settled accuracy.

In~\cite{2019_ICLR_Yu}, Yu \etal  trained a slimmable network with a new variant of BN, namely switchable batch normalization (SBN), for the networks executable at different widths. SBN privatizes BN for different switches of a slimmable network, and each individual BN has independently accumulated feature statistics. SBN can thus be used as a general solution to obtain a good trade-off between accuracy and latency on the fly.
As a complement to BN that normalizes the final summation of the weighted inputs,
Luo \etal~\cite{2020_arxiv_Luo} proposed fine-grained batch normalization (FBN) to build light-weight networks, where FBN normalizes the intermediate state of the summation.

Network quantization is another essential technique in building efficient DNNs. This challenging task can also be tackled using normalization algorithms like BN. For instance,
Banner \etal ~\cite{2018_NeurIPS_Banner} proposed range batch normalization (RBN) for quantized networks,  normalizing activations according to the range of the activation distribution. RBN avoids the sum of squares, square-root and reciprocal operations and is more friendly for low-precise training~\cite{2017_arxiv_Graham}.
Lin \etal~\cite{2020_arxiv_Lin} proposed to quantize BN in model deployment by converting the two floating points affine transformations to a fixed-point operation with shared quantized scale.
Ardakani \etal~\cite{2019_ICLR_Ardakani} employed BN to train binarized/ternarized LSTMs, and achieved state-of-the-art performance in network quantization. Hou \etal ~\cite{2019_NeurIPS_Hou} further studied and compared the quantized LSTMs with WN, LN and BN. They showed that these normalization methods make the gradient invariant to weight scaling, thus alleviating the problem of having a potentially large weight norm increase due to quantization. In~\cite{2019_arxiv_Sari}, Sari \etal analyzed how the centering and scaling operations in BN affect the training of binary neural networks.

\section{Summary and Discussion}
In this paper, we have provided a research landscape for  normalization techniques, covering  methods, analyses and applications.
We believe that our work can provide valuable guidelines for selecting normalization techniques to use  in training DNNs. With the help of these guidelines, it will be possible to design new normalization methods tailored to specific tasks (by the choice of NAP) or improve the trade-off between efficiency and performance (by the choice of NOP).
We leave the following open problems for discussion.

\vspace{0.05in}
\Para{Theoretical Perspective:}
While the practical success of DNNs is indisputable, their theoretical analysis is still limited. Despite the recent progress of deep learning in terms of representation~\cite{2014_NeurIPS_Montufar}, optimization~\cite{2019_arxiv_Sun} and generalization~\cite{2017_ICLR_Zhang}, the networks investigated theoretically are usually different from those used in practice~\cite{2019_ICLR_Yang}. One clear example is that, while normalization techniques are ubiquitously used in the current state-of-the-art architectures, the theoretical analyses  for DNNs usually rule out them.

In fact, the methods commonly used for normalizing activations (\eg, BN, LN) often conflict with current theoretical analyses. For instance, in the representation of DNNs, one important strategy is to analyze the number of linear regions,  where the expressivity of a DNN with rectifier nonlinearity can be quantified by the maximal number of linear regions it can separate its input space into~\cite{2014_NeurIPS_Montufar,2020_ICML_Xionghuan}. However, this generally does not hold if BN/LN are introduced, since they create nonlinearity, causing the the theoretical assumptions to no longer be met. It is thus important to further investigate how BN/LN affect a model's representation capacity.  
As for optimization, most analyses require the input data to be independent, such that the  stochastic/mini-batch gradient is an unbiased estimator of the true gradient over the dataset. However, BN typically does not fit this data-independent assumption, and its optimization usually  depends on the sampling strategy as well as the mini-batch size~\cite{2019_AISTATS_Lian}.  There is thus a need to reformulate the current theoretical framework for optimization when BN is present.

In contrast, normalizing-weights methods do not harm the theoretic analysis of DNNs, and can even attribute to boosting the theoretical results. For example, the Lipschitz constant \wrt a linear layer can be controlled/bounded during training by normalizing the weight with (approximate) orthogonality~\cite{2018_ICLR_Miyato,2020_CVPR_Huang2}, which is an important property for certified defense against adversarial attacks~\cite{2018_NeurIPS_Tsuzuku,2019_ICLR_Anil,2019_ICLR_Qian}, and for theoretically analyzing DNN’s generalization~\cite{2017_NeurIPS_Bartlett,2018_ICLR_Neyshabur}.
However, normalizing weights is still not as effective as normalizing activations when it comes to improving training performance, leaving room for further development.

\vspace{0.05in}
\Para{Applications Perspective:}
 As mentioned previously, normalization methods can be used to `edit' the statistical properties of layer activations, which has been exploited in CV tasks to match particular domain knowledges. However, we note that this mechanism is seldom used in NLP tasks. It would thus be interesting to investigate the correlation between the statistical properties of layer activations and the domain knowledge in NLP, and further improve the performance of the corresponding tasks.
 In addition, There exists an intriguing phenomenon that, while BN/GN work for the CV models, LN is more effective in NLP~\cite{2020_ICML_Shen}.
 Intuitively, BN/GN should work well for NLP tasks, considering that the current state-of-the-art models for CV and NLP tend to be similar (\eg, they both use the convolutional operation and attention) and GN is simply a more general version of LN.
 It is thus important to further investigate whether or not BN/GN can be made to work well for NLP tasks, and, if not, why.

 Another interesting observation is that normalization is not very common in deep reinforcement learning (DRL)~\cite{2020_arxiv_Bhatt}. Considering that certain DRL frameworks (\eg, actor-critic~\cite{2016_ICLR_Lillicrap,2016_ICML_Mnih}) are very similar to  GANs, it should be possible to exploit normalization techniques to improve training in DRL, borrowing ideas from GANs (\eg, normalizing the weights in the discriminator~\cite{2018_ICLR_Miyato,2019_ICLR_Brock,2020_CVPR_Huang2}).


As the key components in DNNs, normalization techniques are links that connect the theory and application of deep learning. We thus believe that these techniques will continue to have a profound impact on the rapidly growing field of deep learning, and we hope that this paper will aid readers in building a comprehensive landscape for their implementation.

\bibliographystyle{IEEEtran}
\bibliography{bib/normalization}

%








\end{document}

%% file: math_commands.tex
\usepackage{amsmath,amsfonts,bm}

\def\eqref#1{equation~\ref{#1}}

\def\1{\bm{1}}

\def\rvh{{\mathbf{h}}}
\def\rvu{{\mathbf{i}}}

\def\rvu{{\mathbf{u}}}

\def\rvw{{\mathbf{w}}}
\def\rvx{{\mathbf{x}}}
\def\rvy{{\mathbf{y}}}

\def\rmH{{\mathbf{H}}}

\def\mD{{\bm{D}}}

\def\mI{{\bm{I}}}

\def\mL{{\bm{L}}}

\def\mV{{\bm{V}}}
\def\mW{{\bm{W}}}
\def\mX{{\bm{X}}}

\DeclareMathAlphabet{\mathsfit}{\encodingdefault}{\sfdefault}{m}{sl}
\SetMathAlphabet{\mathsfit}{bold}{\encodingdefault}{\sfdefault}{bx}{n}
\newcommand{\tens}[1]{\bm{\mathsfit{#1}}}

\def\tW{{\tens{W}}}
\def\tX{{\tens{X}}}

\def\sD{{\mathbb{D}}}

\newcommand{\etens}[1]{\mathsfit{#1}}

\def\etH{{\etens{H}}}

\def\etW{{\etens{W}}}
\def\etX{{\etens{X}}}

\newcommand{\E}{\mathbb{E}}

\newcommand{\R}{\mathbb{R}}

